\documentclass[10pt,twocolumn,letterpaper]{article}

\usepackage{cvpr}              

\usepackage{multirow}
\usepackage{comment}
\usepackage[utf8]{inputenc} 
\usepackage[T1]{fontenc}    
\usepackage{url}            
\usepackage{booktabs}       
\usepackage{amsfonts}       
\usepackage{nicefrac}
\usepackage{microtype}      
\usepackage{xcolor}         
\usepackage{amsmath}
\usepackage{graphicx} 
\usepackage{amssymb}
\usepackage{booktabs}
\usepackage{color}
\usepackage{amsmath,bm}
\usepackage{multirow}
\usepackage{float}
\usepackage{subcaption} 
\usepackage{tabularray}
\usepackage{pifont}
\usepackage{enumitem}
\usepackage{colortbl}
\usepackage{algorithm}
\usepackage{makecell}
\usepackage{newtxtext,newtxmath}

\definecolor{cvprblue}{rgb}{0.21,0.49,0.74}
\usepackage[pagebackref,breaklinks,colorlinks,allcolors=cvprblue]{hyperref}

\def\paperID{2739} 
\def\confName{CVPR}
\def\confYear{2025}

\title{Towards Visual Discrimination and Reasoning of Real-World Physical Dynamics: \\Physics-Grounded Anomaly Detection}

\author{Wenqiao Li$^{1}$\thanks{Equally contribute to this work.}
\quad Yao Gu$^{1}$\footnotemark[1]
\quad Xintao Chen$^{1}$\footnotemark[1]
\quad Xiaohao Xu$^{2}$\footnotemark[2]
\quad Ming Hu$^{3}$
\quad Xiaonan Huang$^{2}$
\quad Yingna Wu$^{1}$\thanks{Corresponding authors}\\
$^{1}$ShanghaiTech University \quad $^{2}$University of Michigan, Ann Arbor \quad $^{3}$Monash University\\}

\begin{document}
\maketitle
\begin{abstract}
Humans detect real-world object anomalies by perceiving, interacting, and reasoning based on object-conditioned physical knowledge. The long-term goal of Industrial Anomaly Detection (IAD) is to enable machines to autonomously replicate this skill. However, current IAD algorithms are largely developed and tested on static, semantically simple datasets, which diverge from real-world scenarios where physical understanding and reasoning are essential.
To bridge this gap, we introduce the Physics Anomaly Detection (Phys-AD) dataset, the first large-scale, real-world, physics-grounded video dataset for industrial anomaly detection. Collected using a real robot arm and motor, Phys-AD provides a diverse set of dynamic, semantically rich scenarios. The dataset includes more than 6400 videos across 22 real-world object categories, interacting with robot arms and motors, and exhibits 47 types of anomalies. Anomaly detection in Phys-AD requires visual reasoning, combining both physical knowledge and video content to determine object abnormality.
We benchmark state-of-the-art anomaly detection methods under three settings: unsupervised AD, weakly-supervised AD, and video-understanding AD, highlighting their limitations in handling physics-grounded anomalies. Additionally, we introduce the Physics Anomaly Explanation (PAEval) metric, designed to assess the ability of visual-language foundation models to not only detect anomalies but also provide accurate explanations for their underlying physical causes. Our project is available at \url{https://guyao2023.github.io/Phys-AD/}.

\end{abstract}    
\section{Introduction}
\label{sec:intro}

\begin{figure}[t!]
  \centering
 \setlength{\abovecaptionskip}{0.1cm}
  \includegraphics[width=0.48\textwidth]{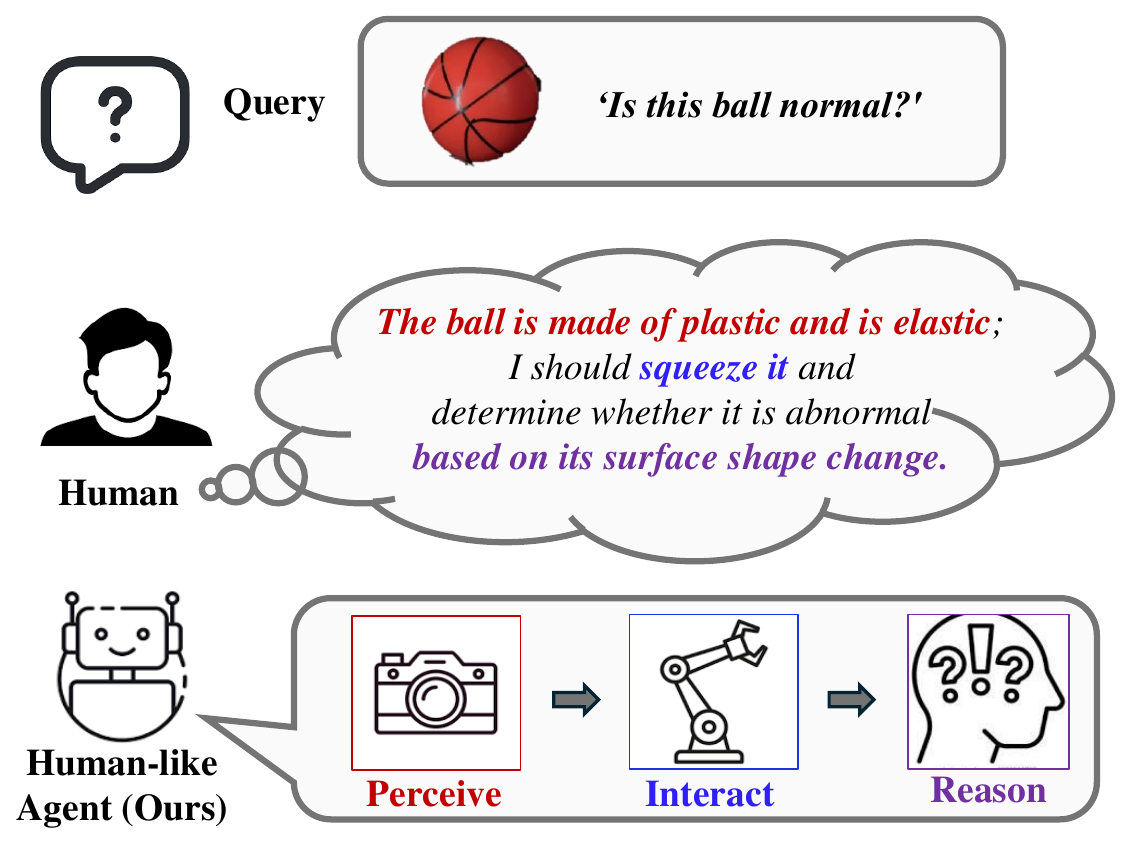}
  \caption{  \textbf{Human-like decision-making process for physics-grounded object anomaly detection.} 
        We illustrate the sequential approach of a human-like agent for evaluating an object’s normality. 
        First, the agent \textcolor{red}{perceives} relevant physical attributes (e.g., plastic and elastic), 
        then \textcolor{blue}{interacts} by performing a physical action (e.g., squeezing), and finally \textcolor[HTML]{660099}{reasons} 
        based on the vision feedback and attributes changes (e.g., surface shape change) to determine whether the object is normal or anomalous. 
        This mirrors a human’s natural process of  reasoning over physics in objects.
   }
  \label{fig:human-like}
\end{figure}

Industrial anomaly detection (IAD) is a critical subfield in computer vision and industrial automation, aiming to identify defects or irregularities in products during manufacturing. As shown in Fig.~\ref{fig:human-like}, the ultimate vision is to create autonomous systems that not only perceive but also interact with and reason about objects to discriminate between anomalies and normal states, integrating complex physical principles to detect anomalies in dynamic, real-world scenarios. For example, a human inspecting a water bottle for anomalies wouldn’t rely solely on visual observation; they might rotate or invert the bottle, using physical interactions and cues, such as noticing a loose cap or an irregular internal flow, to detect issues.

Central to advancing IAD~\cite{cao2024survey} is the availability of high-quality datasets that bridge the gap between academic research and industrial needs. Datasets like MVTec-AD~\cite{MVTec-AD}, MPDD~\cite{9631567}, and VisA~\cite{VisA} have played foundational roles, enabling algorithm development for image-based anomaly detection and bringing IAD to the forefront of computer vision research. While these datasets have significantly advanced single-image anomaly detection, recent datasets, e.g., MVTec-3D~\cite{MVTec-3D}, Real3D~\cite{Real3D}, and Anomaly-ShapeNet~\cite{li2023towards}, have extended IAD to 3D, aligning research more closely with the needs of complex real-world industrial settings.

Yet, as factories increasingly rely on robotic arms and automated systems to perform sophisticated inspections, the limitations of current IAD datasets become apparent. Existing benchmarks focus on static, semantically simple environments, overlooking the physical priors and interactive reasoning required in real-world industrial contexts. This gap highlights a growing need for datasets that not only reflect real-world physical constraints but also challenge models to reason dynamically about anomalies.

To bridge this gap, as shown in Fig.~\ref{fig:1}, we introduce the Physics Anomaly Detection (Phys-AD) dataset, the first large-scale, physics-grounded video dataset for industrial anomaly detection. Phys-AD features over 6,400 videos of 22 categories and 49 objects interacting with robotic arms and motors, capturing 47 types of anomalies that require visual reasoning informed by physical knowledge. The short video clips in the dataset range from 60 to 240 frames in length and are filmed in real-time industrial environment, fully capturing the interaction process between robotic arm or motor and industrial objects. Additionally, to ensure our dataset meets industrial demands and matches the complexity and diversity of the real physical world, we selected industrial objects of different physical qualities, various interaction methods, and anomalies that reflect diverse physical principles and require different reasoning process.
Specifically, we selected 22 object categories spanning across metals, plastics, fluid, amorphous substances and articulated objects with diverse appearances. For interaction, we use mechanical grippers, robotic arms, and motors, incorporating various interaction modes such as pressing, rotating, squeezing, and driving to handle different types of objects.

We benchmark state-of-the-art anomaly detection methods in three key configurations: unsupervised anomaly detection, weakly-supervised anomaly detection, and video-understanding based anomaly detection. Our findings reveal critical gaps in their ability to handle the complexities of physics-grounded scenarios, where anomalies often arise from dynamic, interdependent interactions. To advance the field, we also introduce the Physical Anomaly Explanation (PAEval) metric, designed to assess both detection performance and a model’s ability to explain anomalies by identifying underlying physical causes. Furthermore, our benchmark reveals the fragility of existing methods in tackling these challenging conditions, underscoring the need for approaches that better understand object dynamics and temporal coherence in anomaly detection.

\textbf{Our contributions can be summarized as followings:}
\begin{itemize}
    \item We introduce a novel task of detecting physical-based industrial anomalies in real-world that involves perception, physical and visual reasoning.
    \item We present Phys-AD, the first large-scale, physics-grounded video dataset specifically designed for industrial anomaly detection in real world, containing objects of different physical qualities, multiple interaction methods and various physical reasoning process.
    
    \item We benchmarking the anomaly detection and reasoning performance of popular video AD methods and Visual Language  Foundation Models on the Phys-AD dataset in several settings, establishing a practical and challenging benchmark to promote the development of the physics-related anomaly detection field.
\end{itemize}

\begin{table}[t!]
\centering  \setlength{\abovecaptionskip}{0.1cm}
\caption{\textcolor{black}{\textbf{Comparison of  Phys-AD with existing industrial anomaly detection datasets.} Our Phys-AD dataset is the first to consider \textbf{\textit{complex objects with physical dynamics}}. `Syn', `IR', `D', and `PC' denote Synthetic, Infrared, Depth, and Point Cloud, respectively. \#Anomaly indicates the number of anomaly types.} }
\setlength{\tabcolsep}{1.5pt}
\renewcommand\arraystretch{1.0}
\resizebox{\linewidth}{!}{
\begin{tabular}{@{}l|ccccccc@{}}
\toprule
\multirow{2}{*}{\textbf{Dataset}} &
  \multirow{2}{*}{\textbf{Year}} &
  \multirow{2}{*}{\textbf{Type}} &
  \multirow{2}{*}{\textbf{Modality}} &
  \multicolumn{3}{c}{\textbf{Sample Statistics}} \\ 
  \cmidrule(l){5-7} 
  
                    &      &        &            &Class & \#Anomaly   & \textbf{Physics} &                   \\ \midrule
{MVTec-AD}~\cite{8954181}       & 2019    & Real     & RGB     & 15     & -    & \ding{55} \\
{BTAD}~\cite{Mishra_2021}       & 2021   & Real     & RGB     & 3    & 3    & \ding{55} \\
{MPDD}~\cite{9631567}       & 2021   & Real     & RGB     & 6   &  8   & \ding{55} \\
VisA~\cite{zou2022spotthedifference}         & 2021   & Real     & RGB     & 12    & -   &    \ding{55}\\  
MVTec LOCO-AD~\cite{bergmann2022beyond}         & 2022   & Real     & RGB     & 5    & -   &    \ding{55}\\  

MAD~\cite{zhou2023pad}         & 2023   & Syn+Real     & RGB     & 20    & 3   &    \ding{55}\\ 
LOCO-Annotations~\cite{10710633}    & 2024   & Real     & RGB     & 5    & 5   &    \ding{55}\\  
Real-IAD~\cite{wang2024realiad}        & 2024          & Real     & RGB     & 30      & 8   & \ding{55} \\\midrule

GDXray~\cite{mery2015gdxray}         & 2015   & Real     & X-ray      & 5    & 15   &    \ding{55}\\  
PVEL-AD~\cite{9744494}         & 2023   & Real     & IR     & 1    & 10   &    \ding{55}\\  \midrule

MVTec3D-AD~\cite{Bergmann_2022}        & 2021         & Real     & RGB-D    & 10     & 3-5   & \ding{55} \\
Eyecandies~\cite{bonfiglioli2022eyecandies}        & 2022        & Syn    & RGB-D    & 10     & 3   & \ding{55} \\ \midrule

Real3D-AD~\cite{liu2024real3d}        & 2023        & Real    & PC   & 12    & 2   & \ding{55} \\
Anomaly-ShapeNet~\cite{li2023towards}        & 2024        & Syn   & PC   & 40    & 6  & \ding{55} \\ \midrule

\textbf{Phys-AD (Ours)} & 2024  & Real & RGB     & \textbf{49}     &\textbf{47} & \ding{51}  \\ \bottomrule
\end{tabular}}
\label{tab:1}
\end{table}



\begin{figure*}[t!]
  \centering \setlength{\abovecaptionskip}{0.1cm}
  \includegraphics[width=1.0\textwidth]{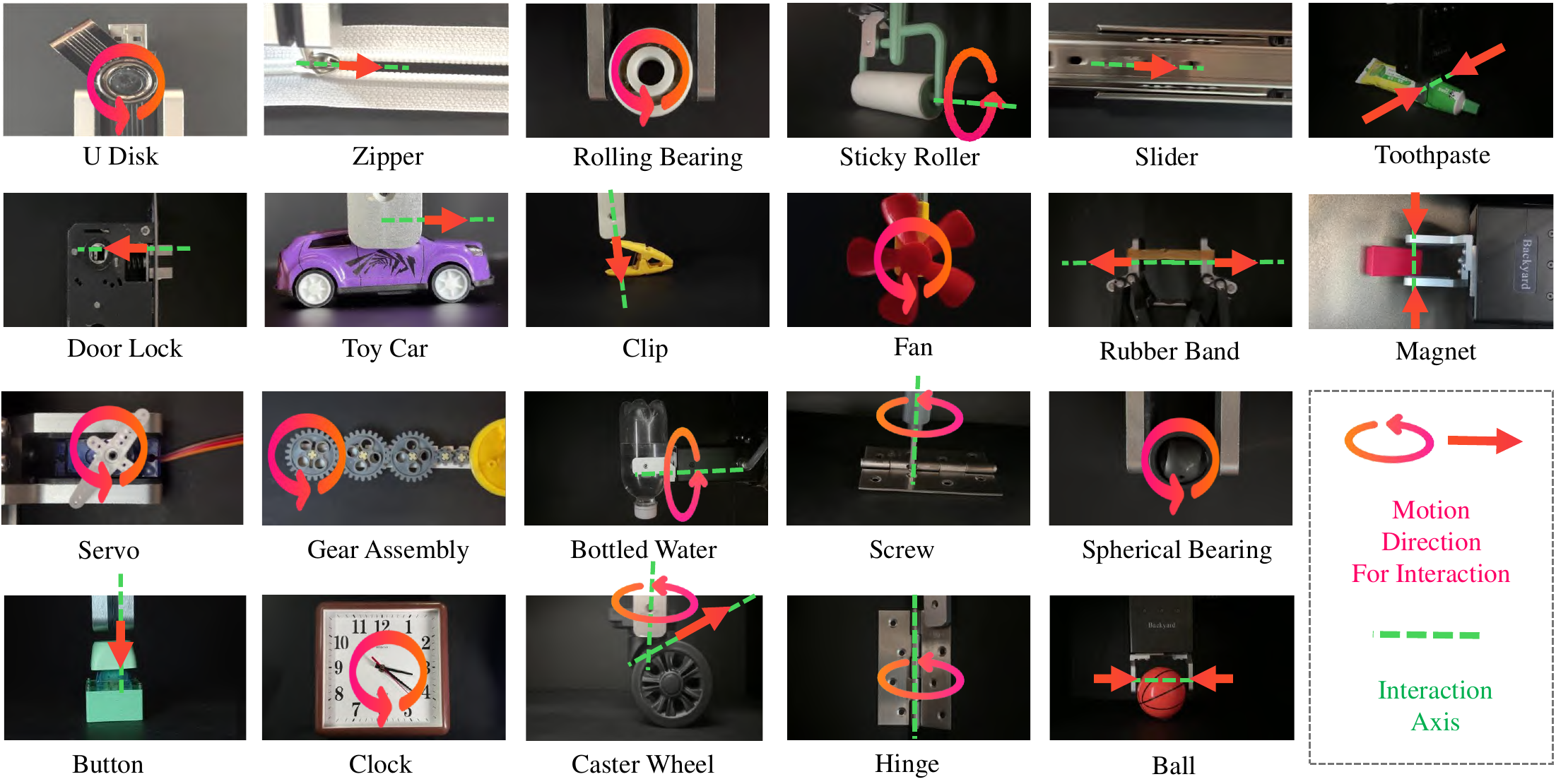}
  \caption{\textcolor{black}{        \textbf{Interactions for understanding implicit physical laws in the Phys-AD dataset.} 
        We showcase various object interactions from the Phys-AD dataset, 
        where different actions (indicated by motion directions) are used to explore 
        and reason about the underlying physical properties and behaviors of each object.
        The colored arrows indicate the interaction directions and axes, highlighting 
        how physical interactions reveal the implicit physics governing each object.}}
  \label{fig:3}\vspace{-2mm}
\end{figure*}

\begin{figure*}[t!]
  \centering \setlength{\abovecaptionskip}{0.1cm}
  \includegraphics[width=1.0\textwidth]{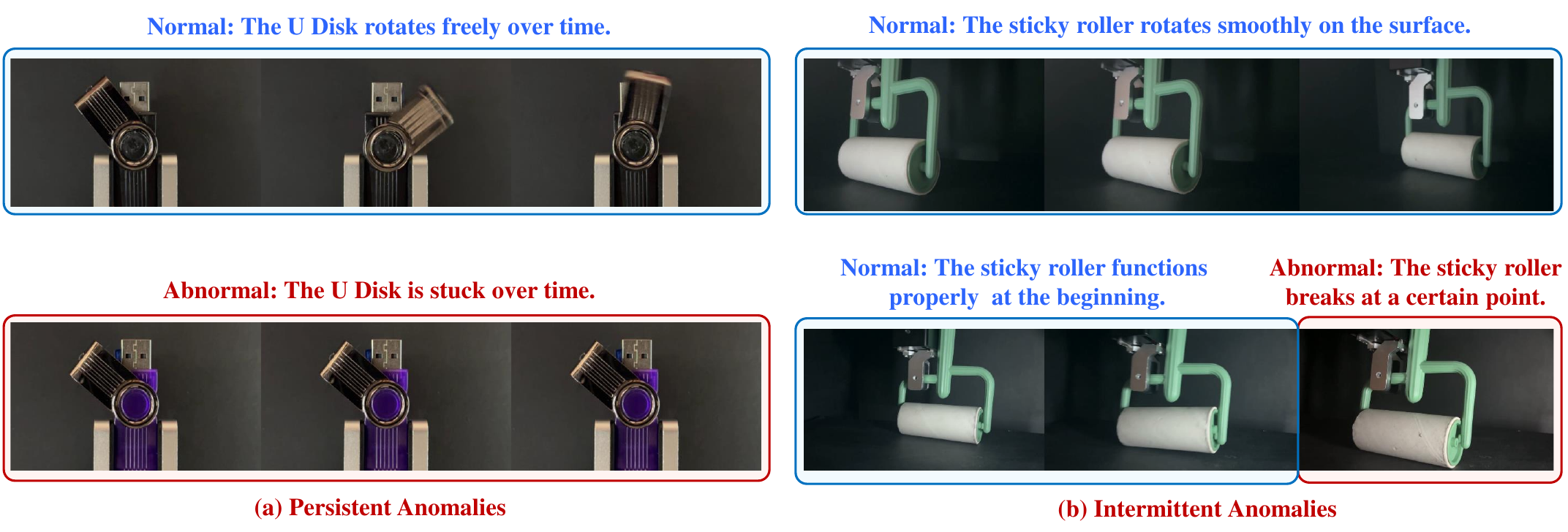}
  \caption{\textcolor{black}{        \textbf{Categorization of anomalies based on persistence in the Phys-AD dataset.} We show examples of normal and abnormal functioning in common objects, divided into two anomaly types: persistent and intermittent. (\textbf{a}) Persistent anomalies, such as continuous obstruction in the U Disk or permanent malfunction of the Sticky Roller, are visible throughout the operation. (\textbf{b}) In contrast, intermittent anomalies, like occasional jamming of the U Disk or breakage in the Sticky Roller after initial operation, only appear at specific points in time. This classification provides insight into both constant and sporadic failures in object interactions.}}
  \label{fig:4}
\end{figure*}

\section{Related work}\label{sec:related_work}

\textbf{Industrial anomaly detection datasets.}  
\textcolor{black}Existing datasets primarily focus on static, semantically simple scenarios, deviate significantly from real world where physical understanding and reasoning are essential. Datasets like MVTec-AD~\cite{8954181}, BTAD~\cite{Mishra_2021}, MPDD~\cite{9631567}, and VisA~\cite{zou2022spotthedifference} focus on surface level image anomaly detection with one single-view RGB image, limiting their effectiveness in capturing holistic object structures. MVTec-LOCO-AD~\cite{bergmann2022beyond} ,focusing on structure and local information in industries , is limited by its relatively simple and constrained data content. While MVTec3D-AD~\cite{Bergmann_2022} and Eyecandies~\cite{bonfiglioli2022eyecandies} incorporate depth data, they remain static and single-view image anomaly detection, neglecting object level information. To explore object level anomaly detection, multi-view IAD datasets like MAD~\cite{zhou2023pad} and Real-IAD~\cite{wang2024realiad}, point cloud IAD datasets like Real3D-AD~\cite{liu2024real3d} and Anomaly-ShapeNet~\cite{li2023towards}, offering richer visual and geometric cues, but are still limited to static objects without dynamic interaction or reasoning. In summary, \textbf{\textit{current industrial anomaly detection datasets focus on relatively simple and static anomaly detection scenarios, lacking complex physical rules, dynamic interactions, and visual reasoning requirements.}} Therefore, existing IAD datasets generally applied to limited industrial scenarios and there is no IAD dataset could meet the demands of detecting complex anomalies in real world which need physical priors and reasoning. 

\vspace{1mm}
\noindent\textbf{Video anomaly detection.} \textcolor{black}{Deep learning methods~\cite{ConvAE,yang2023video,UMIL,zhang2024holmes} now dominate Video Anomaly Detection (VAD), categorized into unsupervised, weakly-supervised, and fully-supervised approaches. Unsupervised methods learn normal patterns via reconstruction~\cite{ConvAE,xu2017detecting,gong2019memorizing}, prediction~\cite{framepred}, or hybrids~\cite{liu2021hybrid}, while some methods~\cite{zaheer2022generative,thakare2023dyannet} train with both unlabeled normal and abnormal data. Weakly-supervised methods\cite{GCN,mist,xdviolence,zhang2024glancevad} use video-level or glance-based annotations, and fully-supervised methods\cite{liu2019exploring,landi2019anomaly} remain rare due to costly frame-level labeling.
Visual-language models like CLIP~\cite{radford2021learning} have recently been applied to enhance anomaly detection~\cite{PromptEnhanced,cliptsa,vadclip}, focusing on semantic anomalies. Open-vocabulary VAD~\cite{wu2023open} and prompt-based anomaly scoring~\cite{zanella2024harnessing} leverage LLMs~\cite{zhang2024holmes,xu2024customizing}, but performance relies heavily on the base models, often lacking domain-specific tuning.
\textbf{\textit{However, existing video anomaly detection algorithms lack the capability to handle complex industrial anomaly detection scenarios and understand physical rules.}} This gap highlights the need for models that can capture dynamic behaviors and physical laws in industrial environments, as addressed by Phys-AD, which targets industrial anomalies in objects of various physical properties.}


\vspace{1mm}
\noindent\textbf{Visual reasoning.}
Visual reasoning is a critical task in computer vision, aiming to enable machines to interpret perceptual information like humans. Several visual reasoning tasks have been proposed to evaluate reasoning capabilities, including Visual Question Answering (VQA), 2D puzzles, and physical dynamics prediction.
In VQA, agents are required to combine natural language and visual cues to answer questions~\cite{Antol_2015_ICCV,Goyal_2017_CVPR,Hudson_2019_CVPR,Johnson_2017_CVPR}. For 2D puzzles, tasks involve discovering relationships among visual elements and making inferences~\cite{li2011comparing,zerroug2022benchmark,zhang2019raven,jiang2022bongard,Nie2020Bongard}.
Physical dynamics prediction tasks require machines to perceive and reason about physical interactions~\cite{baradel2019cophy,duan2021space,duan2022pip,janny2022FilteredCoPhy}.
In contrast to these work, \textbf{\textit{we introduce the first benchmark featuring real-world industrial objects with dynamic physical properties, focusing on distinguishing diverse dynamics through vision}}.

\begin{figure}[t!]
  \centering  \setlength{\abovecaptionskip}{0.1cm}
  \includegraphics[width=0.48\textwidth]{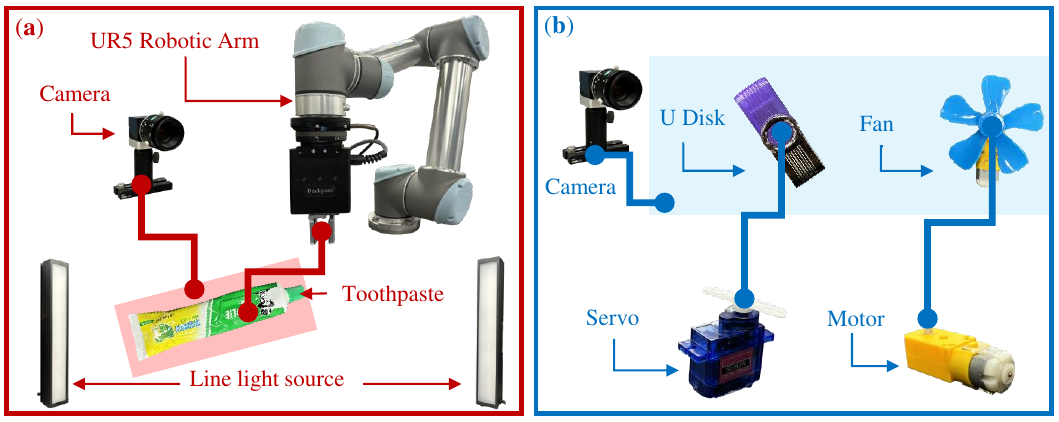}
  \caption{\textcolor{black}{\textbf{Data collection pipeline for the Phys-AD dataset.} (\textbf{a}) Manipulation of a toothpaste tube using a UR5 robotic arm. (\textbf{b}) Manipulation of a U Disk and fan via servo and motor.}}
  \label{fig:5}
\end{figure}

\section{Dataset: Phys-AD}
\label{construction}

\subsection{Object Preparation and Interaction Selection}
We selected 22 categories spanning across various materials, including metals, plastics, amorphous substances and articulated objects, with diverse shapes, sizes, and physical properties. For the physical properties of different objects, we correspondingly select different interaction methods like push, rotate, pull with robotics arms and motors. For instance, we use robotics arms to grab and extruding the deformable objects like basketball to determine whether there are any elastic anomalies or surface defects by the morphological change of the objects. To make our dataset more practical and challenging, we introduce 47 distinct defect types, some of which just rely on single frame content and physical rules for anomaly reasoning, while the other part needs to combine the content of the whole video and physical priors together to judge whether the object is abnormal or not. Different objects information with their corresponding interaction methods and anomalies are listed in Table~\ref{tab:2}. Fig \ref{fig:3} shows the interactions for understanding implicit physical laws in the Phys-AD dataset. Fig \ref{fig:4} provides two representative examples: one where a U Disk requires analyzing the entire video content to determine if there is an anomaly, and another where a sticky roller only requires a few frames from the video to judge whether the object is abnormal.

\begin{table}[t!]
\centering  \setlength{\abovecaptionskip}{0.1cm}
\caption{\textbf{Interaction methods and anomalies of the object categories in our Phys-AD dataset.} Motor. means one object driven by motor. Multi. means one object with multiple anomalies.}
\setlength{\tabcolsep}{4pt} 
\renewcommand\arraystretch{1.0} 
\resizebox{\linewidth}{!}{
\begin{tabular}{l|c|l}
\toprule
\textbf{Category} & \textbf{Interaction} & \textbf{Anomaly Types} \\ \midrule
\makecell[l]{Car}              & Drag, Slide     & Different wheels stuck                 \\ \midrule
\makecell[l]{Fan}              & Motor., Rotate & Stuck, Uneven Rotation, Vibration      \\ \midrule
\makecell[l]{Rolling\\Bearing}  & Motor., Rotate     & Lack of friction             \\ \midrule
\makecell[l]{Spherical\\Bearing} & Grab, Rotate       & Internal block              \\ \midrule
\makecell[l]{Servo}            & Grab, Motor., Rotate& Angle restricted, Vibration, No calibration \\ \midrule
\makecell[l]{Clip}             & Press              & Unable to press down, Unable to rebound          \\ \midrule
\makecell[l]{U Disk}              & Grab, Motor., Rotate & Cover jam                   \\ \midrule
\makecell[l]{Hinge}            & Grab, Rotate       & Angle restricted, Shaft off  \\ \midrule
\makecell[l]{Sticky\\Roller}    & Grab, Pull         & Detach, Unable to rotate \\ \midrule
\makecell[l]{Caster\\Wheel}     & Slide              & Axle axis stuck, Swivel axis stuck \\ \midrule
\makecell[l]{Screw}            & Press, Rotate      & Loosen, Unable to insert \\ \midrule
\makecell[l]{Lock}             & Motor., Rotate     & Latch jam, Loose    \\ \midrule
\makecell[l]{Gear}             & Motor., Rotate     & Stuck, Not meshed, Multi. \\ \midrule
\makecell[l]{Clock}            & Motor., Rotate     & Pointer stops, Vibration \\ \midrule
\makecell[l]{Slide}            & Grab, Slide        & Detach, Shedding, Jam \\ \midrule
\makecell[l]{Zipper}           & Grab, Close        & Stuck, Unable to close      \\ \midrule
\makecell[l]{Button}           & Press              & No light, Unable to press down, Unable to rebound, Multi. \\ \midrule
\makecell[l]{Liquid}           & Grab, Shake        & Water out, Foreign body \\ \midrule
\makecell[l]{Rubber\\Band}      & Stretch            & Crack                   \\ \midrule
\makecell[l]{Ball}             & Pinch, Press       & Insufficient gas, Leakage \\ \midrule
\makecell[l]{Magnet}           & Grab, Press, Move  & Degaussing, Shell detached \\ \midrule
\makecell[l]{Toothpaste}       & Pinch, Press       & Leakage                        \\ 
\bottomrule
\end{tabular}
}
\label{tab:2}\vspace{-2mm}
\end{table}

\begin{table}[t!]
\centering  \setlength{\abovecaptionskip}{0.1cm}
\caption{\textbf{Phys-AD dataset statistics.} We denote the total number of frames and videos for each category as \#Images and \#Videos. Note that the \textit{Train} split does not contain anomaly data.}
\setlength{\tabcolsep}{1mm}
\renewcommand\arraystretch{1.0}
\resizebox{\linewidth}{!}{
\begin{tabular}{l|cc|cccc|cc}
\toprule
\textbf{Category} & \makecell{\textbf{Train} \\ \textbf{\#Frames}} & \makecell{\textbf{Train} \\ \textbf{\#Videos}} & \makecell{\textbf{Test} \\ \textbf{\#Frames}} & \makecell{\textbf{Test} \\ \textbf{\#Videos}} & \makecell{\textbf{Test} \\ \textbf{\#Videos Nor.}} & \makecell{\textbf{Test} \\ \textbf{\#Videos Ab.}} & \makecell{\textbf{Obj.} \\ \textbf{Types}} & \makecell{\textbf{Def.} \\ \textbf{Types}} \\ \midrule
Car              &18,000   & 300   & 36,000  & 600   & 150   & 450   & 5  & 3 \\ 
Fan              &32,400   & 180  & 64,800  & 360  & 90    & 270   & 3  & 3 \\ 
Rolling Bearing  & 3,600   & 60   & 3,600   & 60   & 30    & 30    & 1  & 1 \\ 
Spherical Bearing& 3,600   & 60   & 3,600   & 60   & 30    & 30    & 1  & 1 \\ 
Servo            & 14,400  & 120  & 28,800  & 240  & 60    & 180   & 1  & 3 \\ 
Clip             &28,800   & 240  & 43,200  & 360  & 120   & 240   & 4  & 2 \\ 
U Disk              &14,400   & 240   &14,400  & 240   & 120   & 120   & 4  & 1 \\ 
Hinge            &3,600    & 30  & 7,200   & 60  & 15    & 45    & 1  & 2 \\ 
Sticky Roller    & 5,400   & 30  & 8,100   & 45  & 15    & 30    & 1  & 2 \\ 
Caster Wheel     &5,400    & 30  & 10,800   & 60  & 15    & 45    & 1  & 2 \\ 
Screw            &5,400    & 30  & 8,100   &45  & 15    & 30    & 1  & 2 \\ 
Lock             &7,200   & 120   & 10,800  & 180   & 60    & 120   & 1  & 2 \\ 
Gear             &21,600   & 180  & 54,000  & 450  & 90    & 360   & 3  & 4 \\ 
Clock            &27,000   & 150  & 40,320  & 224  & 73    & 151   & 5  & 2 \\ 
Slide            &7,200    & 60  & 18,000  & 150  & 30    & 120   & 1  & 3 \\ 
Zipper           &14,400  & 120  & 21,600  & 180  & 60    & 120   & 2  & 2 \\ 
Button           &21,600   & 120  & 54,000  & 300  & 60    & 240   & 4  & 4 \\ 
Liquid           & 5,400   & 30  & 8,100   & 45  & 15    & 30    & 1  & 2 \\ 
Rubber Band      &10,800    & 60  & 10,800   & 60  & 30    & 30    & 1  & 1 \\ 
Ball             &21,600   & 90  & 32,400 & 135  & 45    & 30    & 3  & 2 \\ 
Magnet           &10,800    & 60  & 16,200   & 90  & 30    & 60    & 2  & 2 \\ 
Toothpaste       &16,200    & 90  & 16,200   & 90  & 45    & 45    & 3  & 1 \\ \midrule
\textbf{Total}   &298,800  & 2400 & 511,020 & 4,034 & 1,198  & 2,836  & 49 & 47 \\ \bottomrule
\end{tabular}
}
\label{tab:3}\vspace{-2mm}
\end{table}

\subsection{Data Collection and Processing} 

\noindent \textbf{Data collection pipeline.} Most of the data collection for the Phys-AD dataset is driven by manipulation-guided video sequences, captured using a UR5 robotic arm equipped with an RGB camera (see Fig.~\ref{fig:5}a). In order to reproduce real industrial scenes, we also include adjustable light sources in our data capture process.  The RGB Camera, with a 1080p resolution and a frame rate of 60 FPS, providing high-quality video sequences. Some kinds of the objects like U Disk that are not suitable for the manipulation of robotics arms, are driven by the motor or servo (Fig.~\ref{fig:5}b). After the data collection, we used video editing software to crop out irrelevant frames and retain the complete interaction process.

\subsection{Data Statistics}
\noindent \textbf{Dataset sample distribution.} Table~\ref{tab:3} provides a detailed breakdown of the Phys-AD dataset, which includes information on category, the number of training videos and frames, the number of testing videos and frames, the distribution of normal and abnormal samples in the test set, and the number of object and defect types. The length of the video clips in the dataset ranges from 60 to 240 frames, ensuring to fully capture the interaction process in a short time. The dataset contains 2400 training videos and 4034 test videos spreading across 22 categories, 49 object types, and 47 defect types. The frames are extracted from the videos at a rate of 60 FPS. The test set includes 1,198 normal samples and 2,836 abnormal samples. Overall, the Phys-AD dataset covers a wide range of categories and contains a large scale of data, which helps to train more robust anomaly detection models and provides a reasonable evaluation setting.

\vspace{1mm}
\noindent \textbf{Interaction and defect types across categories.} Table~\ref{tab:2} details the defect types and corresponding interaction methods for each category in the Phys-AD dataset. Covering 22 categories and including 47 defect types, the dataset features various interactions (e.g., Rotate, Grab). Importantly, interaction methods or defect types with identical names can differ across categories—for example, "rotation" for a fan describes the motor-driven movement of its blades, while for a hinge it refers to the turning of its pages around the shaft under robotic control. Furthermore, to detect anomalies with complex physical properties in a human-like manner, we sometimes combine multiple interactions. For instance, when inspecting a screw, we first press it into the hole and then use a robotic arm to rotate it further. Overall, Phys-AD provides a diverse, physics-grounded anomaly detection framework that fosters the development of advanced methods for tackling real-world challenges.

\begin{figure}[t!]
  \centering  \setlength{\abovecaptionskip}{0.1cm}
  \includegraphics[width=0.48\textwidth]{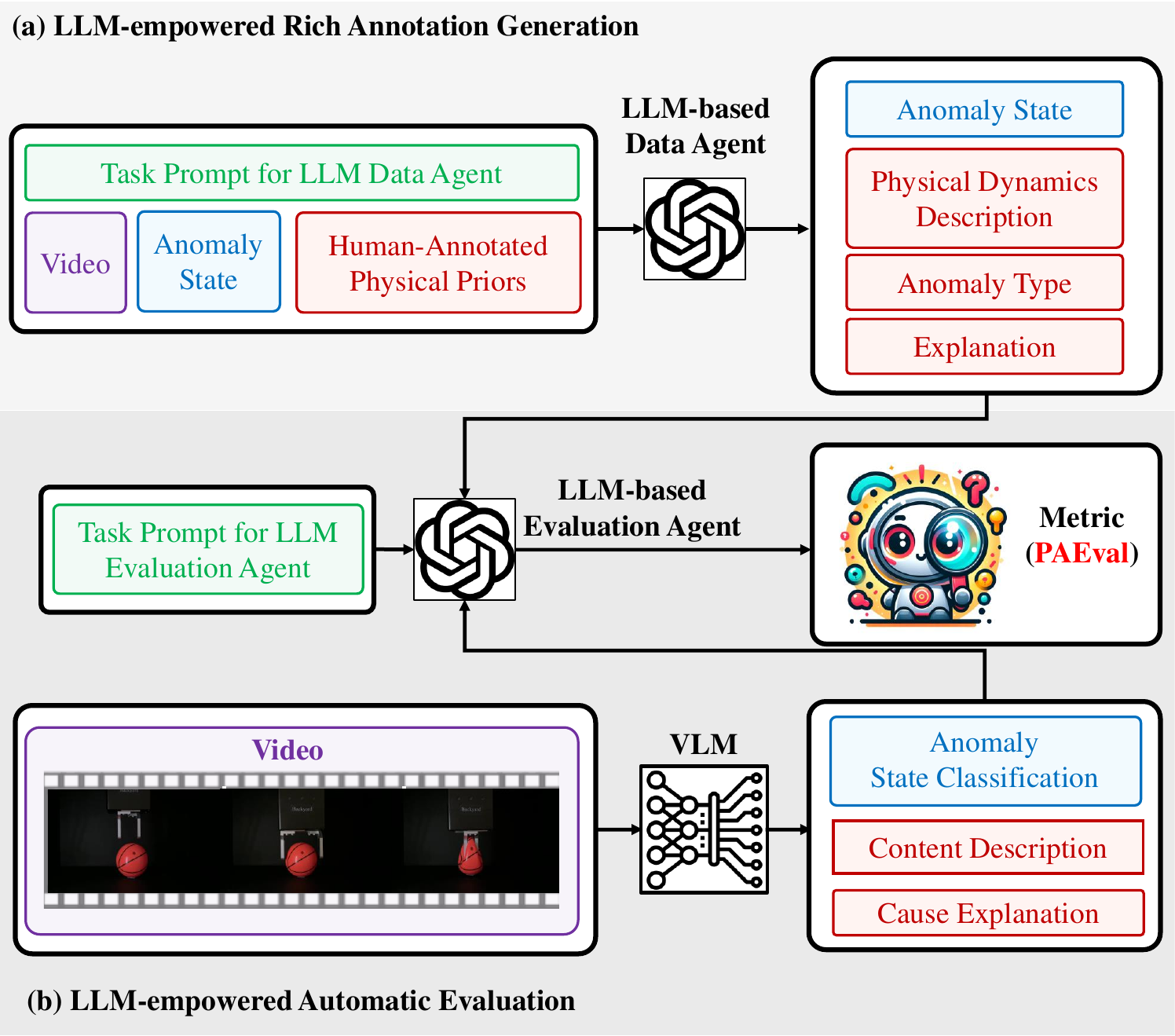}
  \caption{\textcolor{black}{\textbf{PhysAD-Agent: A Large Language Model (LLM)-powered system for physics anomaly detection label augmentation and automatic evaluation. }This agent framework consists of two main components: (\textbf{a}) \textbf{\textit{Rich Annotation Generation}}, where an LLM-based data agent generates detailed anomaly annotations based on video, anomaly state, prompt, and human-annotated physical priors, and (\textbf{b}) \textbf{\textit{Automatic Evaluation}}, where an LLM-based evaluation agent assesses model predictions to calculate the Physics Anomaly Explanation (PAEval) metric. }}
  \label{fig:device}
\end{figure}

\vspace{1mm}
\noindent \textbf{Labels.} For the test set videos, we provide two types of labels. First, there are the common video-level labels: all anomalous videos in the test set are labeled as 1, while normal videos are labeled as 0. For evaluating visual-language models, we also provide text labels. Specifically, for each type of anomaly in the test set, we manually design a textual description label. This label includes both a description of the video content and a physical explanation of the reason for the anomaly. To ensure diversity in our text labels, we also use ChatGPT-4o for text augmentation. For further details on the labeling process, please refer to Fig \ref{fig:device}a.

\begin{table*}[t!]
\centering
\caption{\textcolor{black}{\textbf{Video-level AUROC ($\uparrow$) result of 22 categories on Phys-AD dataset.} We include Unsupervised, Weakly-supervised and Video-understanding methods. The best per-category result for each class of methods is highlighted in \textbf{bold}.}‘ZS ImgB’,‘V-ChatGPT',‘V-LLaMA',‘V-LLaVA' denote ZS ImageBind,Video-ChatGPT,Video-LLaMA and Video-LLaVA.}
\vspace{-5pt}
\centering\setlength{\tabcolsep}{1mm}
\resizebox{\textwidth}{!}{
\begin{tabular}{l|ccccc|ccc|cccccc}
\toprule
\multirow{2}{*}{\textbf{Category.}} & 
    \multicolumn{5}{c|}{\textbf{Unsupervised}} & 
    \multicolumn{3}{c|}{\textbf{Weakly-supervised}} &
    \multicolumn{6}{c}{\textbf{Video-understanding}} \\
    \cmidrule(l){2-6} 
    \cmidrule(l){7-9} 
    \cmidrule(l){10-15}
& \makecell[c]{MPN\\\cite{lv2021learning}} & \makecell[c]{MemAE\\\cite{gong2019memorizing}} & \makecell[c]{MNAD.p\\\cite{park2020learning}} & \makecell[c]{MNAD.r\\\cite{park2020learning}} & \makecell[c]{SVM\\\cite{sultani2018real}} & \makecell[c]{VADClip\\\cite{vadclip}} & \makecell[c]{S3R\\\cite{S3R}} & \makecell[c]{MGFN\\\cite{chen2023mgfn}} & \makecell[c]{LAVAD\\\cite{zanella2024harnessing}} & \makecell[c]{ZS Clip\\\cite{radford2021learning}} & \makecell[c]{ZS ImgB\\\cite{girdhar2023imagebind}} & \makecell[c]{V-ChatGPT\\\cite{maaz2023video}} & \makecell[c]{V-LLaMA\\\cite{zhang2023video}} & \makecell[c]{V-LLaVA\\\cite{lin2023video}} \\ \midrule

Car               &0.229  &0.523  &0.492  &\textbf{0.944}  &0.587    &0.581  &\textbf{0.606}  &0.571       &0.557 &0.500 &0.500 &0.500 &\textbf{0.678}&0.522  \\ 
Fan         &\textbf{0.811}  &0.371  &0.810  &0.542  &0.500    &0.624  &\textbf{0.640}  &0.542       &0.510 &0.500 &0.500 &0.549 &0.592&\textbf{0.611}  \\   
Rolling Bearing   &0.353  &0.044  &0.352  &0.800  &\textbf{0.933}    &0.589  &0.601  &\textbf{0.680}       &\textbf{0.532} &0.500 &0.500 &0.300 &0.500&0.500  \\ 
Spherical Bearing &0.113  &0.092  &\textbf{0.962}  &0.813  &0.650    &0.500  &\textbf{0.682}  &0.583       &0.435 &0.500 &0.500 &0.450 &\textbf{0.550}&0.500  \\ 
Servo             &0.364  &0.445  &\textbf{0.975}  &0.878  &0.500    &0.518  &\textbf{0.592}  &0.556       &0.502 &0.500 &0.500 &0.506 &\textbf{0.683}&0.464  \\  
Clip              &0.535  &0.443  &\textbf{0.630}  &0.333  &0.500    &0.412  &0.561  &\textbf{0.563}       &0.516 &0.500 &0.500 &\textbf{0.669} &0.556&0.458  \\  
U Disk               &0.240  &0.617  &0.609  &\textbf{0.940}  &0.500    &0.530  &0.549  &\textbf{0.570}       &0.513 &0.500 &0.500 &0.565 &\textbf{0.575}&0.500  \\   
Hinge             &0.769  &0.870  &0.705  &\textbf{0.895}  &0.500    &\textbf{0.737}  &0.561  &0.550       &0.564 &0.500 &0.500 &0.500 &\textbf{0.692}&0.500  \\   
Sticky Roller     &\textbf{0.967}  &\textbf{0.967}  &0.451  &0.936  &0.500    &0.542  &\textbf{0.835}  &0.669       &0.266 &0.500 &0.500 &0.450 &\textbf{0.544}&0.467  \\   
Caster Wheel      &0.364  &\textbf{0.523}  &0.271  &0.508  &0.500    &0.587  &\textbf{0.676}  &\textbf{0.676}       &0.615 &0.500 &0.500 &0.444 &\textbf{0.642}&0.500  \\   
Screw             &0.522  &0.567  &\textbf{0.680}  &0.547  &0.500    &0.500  &\textbf{0.657}  &0.541       &\textbf{0.688} &0.500 &0.500 &0.472 &0.256&0.550  \\   
Lock              &0.563  &0.597  &0.430  &0.641  &\textbf{0.733}    &0.452  &0.523  &\textbf{0.626}       &0.341 &\textbf{0.500} &\textbf{0.500} &0.279 &0.494&\textbf{0.500}  \\   
Gear              &0.529  &0.510  &0.652  &\textbf{0.694}  &0.500    &0.519  &\textbf{0.580}  &0.544       &\textbf{0.603} &0.500 &0.500 &0.544 &\textbf{0.603}&0.517  \\   
Clock             &0.340  &0.542  &0.395  &\textbf{0.587}  &0.500    &0.500  &\textbf{0.549}  &0.509       &0.500 &0.500 &0.500 &\textbf{0.501} &0.360&0.500  \\   
Slide             &0.517  &\textbf{0.962}  &0.917  &0.784  &0.500    &0.531  &\textbf{0.611}  &0.568       &0.425 &0.500 &0.500 &0.562 &\textbf{0.567}&0.179  \\  
Zipper            &0.815  &0.592  &\textbf{0.829}  &0.421  &0.500    &0.504  &\textbf{0.636}  &0.633       &\textbf{0.535} &0.500 &0.500 &0.547 &0.489&0.500  \\   
Button            &\textbf{0.853}  &0.365  &0.660  &0.568  &0.500    &\textbf{0.627}  &0.515  &0.566       &0.439 &0.500 &0.500 &0.515 &\textbf{0.517}&0.360  \\   
Liquid            &0.184  &0.700  &0.671  &\textbf{0.831}  &0.500    &0.542  &0.595  &\textbf{0.793}       &\textbf{0.504} &0.500 &0.500 &0.410 &0.278&0.217  \\  
Rubber Band       &0.374  &0.368  &0.366  &0.307  &\textbf{0.567}    &0.482  &\textbf{0.623}  &0.604       &0.511 &0.500 &0.500 &\textbf{0.517} &0.450&0.450  \\  
Ball              &0.543  &0.383  &\textbf{0.728}  &0.687  &0.500    &0.500  &\textbf{0.667}  &0.567       &0.603 &0.500 &0.500 &0.562 &\textbf{0.636}&0.533  \\  
Magnet            &\textbf{0.671}  &0.464  &0.691  &0.438  &0.500    &0.500  &0.548  &\textbf{0.719}       &0.502 &0.500 &0.500 &\textbf{0.683} &0.300&0.400  \\   
Toothpaste        &0.587  &\textbf{0.889}  &0.520  &0.631  &0.500    &0.500  &0.686  &\textbf{0.711}       &\textbf{0.562} &0.500 &0.500 &0.376 &0.550&0.467  \\  \midrule
Average           &0.511  &0.538  &0.627  &\textbf{0.669}  &0.544    &0.535  &\textbf{0.613}  &0.606       &0.510 &0.500 &0.500 &0.496 &\textbf{0.523}&0.463  \\  \bottomrule
\end{tabular}
}
\label{tab:4}\vspace{-2mm}
\end{table*}

\section{Phys-AD Benchmark}
\label{Benchmark}
\subsection{Problem Definition and Challenges}
We establish unsupervised and weakly supervised AD settings for Phys-AD, using unsupervised as the default in our problem definition.\\
We formulate our Phys-AD setting as two steps.\\
\noindent\textbf{Step 1: Rules deduction} Given a set of training objects $\mathcal{T} = \left\{ t_i \right\}_{i=1}^{N}$ from category $c_i$, we first use robotics arms and motors to interact with $\mathcal{T}$ in corresponding methods ${I_i}$, and we get the interaction process in video format as ${V_i}$. Then, we feed the video sequence ${V_i}$ and the corresponding category's physical prior ${P_i}$ together to the deduction function ${f}$ to obtain the normal rules ${r_i}$. After rules deduction for all the categories, we get the rules bank $\mathcal{M} = \left\{ r_i \right\}_{i=1}^{N}$. Step 1 can be formulated as below:
\begin{equation}
    \mathcal{T} \Theta I_i = V_i
\end{equation}
\begin{equation}  
    f(V_i| P_i) = r_i
\end{equation}
\begin{equation}
    \mathcal{M} = \left\{ r_i \right\}_{i=1}^{N}
\end{equation}
$\Theta$ represents interaction between objects and robotics arms. $N$ is the total number of categories.\\
\noindent\textbf{Step 2: Anomaly reasoning} During test time, we first use robotics arms and motors to interact with $\mathcal{T}$ from the test set in corresponding methods ${I_i}$, and we get the interaction process in video format as ${V_i}$. Then, we feed the video sequence ${V_i}$ and rules ${r_i}$ from step 1 to the reasoner ${R}$ to predict the object's anomaly score $S$. Step 2 can be formulated as:
\begin{equation}
    \mathcal{T} \Theta I_i = V_i
\end{equation}
\begin{equation}  
    R(V_i| r_i) = S, r_i \in \mathcal{M}
\end{equation}
\noindent\textbf{Challenges.} Key challenges include combining video information with objects' physical priors to deduct normal rules, capturing long-term temporal dependencies and fine-grained frame-level information, ensuring robust reasoning for anomaly prediction, and generalizing across diverse anomaly patterns. Addressing these challenges is essential for advancing anomaly detection in real-world environment.

\subsection{Evaluation Settings}
\noindent\textbf{Unsupervised AD.} Unsupervised AD is the most common IAD setting for existing IAD datasets and algorithms. Under unsupervised AD setting, the training set contains only normal video data, and the algorithm needs to capture the normal distribution of the data from the training set. In the test set, both normal and abnormal data are included, and the algorithm must distinguish between normal and abnormal data based on the distribution learned during training. \\
\noindent\textbf{Weakly-supervised AD.} Our dataset is in the form of videos for complex anomaly detection in industrial scenarios. In the video context, it is inevitable that we need to discuss weakly supervised anomaly detection. Under weakly-supervised Phys-AD setting, 2\textasciitilde4 video-level labeled anomaly samples are sampled from all possible anomaly classes in the test set in our Phys-AD dataset. These sampled anomalies are then removed from the test set. It's worth noting that we only provide video level label in our test set. This is because most anomalies in our test set require reasoning based on the entire video content and physical knowledge of the objects.\\
\noindent\textbf{Video-Understanding AD.} Video-Understanding models are another potential solution to our Phys-AD setting. Similar to unsupervised setting, we only provide normal videos during training. The visual language models need to provide explicit normal rules in language format during training. In the test phase, these video-understanding AD methods have to truly understand the videos and predict the anomalies based on the language rules withdrawn from the training phase.\\
\subsection{Evaluation Metrics}
\noindent\textbf{Physics anomaly accuracy metrics.} We use the Area Under the Receiver Operating Characteristic Curve (AUROC) to evaluate video-level anomaly detection performance. We also report the average precision (AP), \textit{i.e.} the area under the video-level precision-recall curve and the acc (ACCURACY), following previous works \cite{wang2023multimodal,xdviolence}.

\vspace{1mm}
\noindent\textbf{Physics Anomaly Explanation (PAEval) metric.} Directly utilizing 
 existing VLMs \cite{zhang2023video}\cite{maaz2023video} to understand the videos in our dataset and detect the anomalies is a potential solution to our challenge. The key point is: \textbf{\textit{Can existing VLMs truly understand physical rules and reason in a right way?}} Specifically designed for video-understanding methods, we introduce a new metric named Physics Anomaly Explanation (PAEval) metric. To be more specific, PAEval evaluate the anomaly detection performance of algorithms based on VLMs from three different perspectives: classification, description, and explanation. Classification refers to the traditional anomaly detection metrics like AUROC, etc. Inspired by the works \cite{du2024uncovering,zhang2024holmes}, PAEval also introduces two additional evaluation metrics: description and physical explanation. Description refers to the model's ability to correctly describe the content of the video, used to assess whether the model has the capability to describe physical phenomena. Explanation refers to the model's ability to correctly explain the physical causes of anomalies in the video, used to evaluate the VLM's reasoning ability. To provide labels for description and explanation metrics, we manually labeled each type of defects from all categories and performed data augmentation by ChatGPT to ensure the robustness of the detection. The whole pipeline of PAEval metric is depicted in Fig \ref{fig:device}.




\section{Benchmarking Results}

\subsection{Benchmark Methods Selection} 
For the Phys-AD setting, we select popular and reproducible video anomaly detection methods across unsupervised, weakly-supervised, and video-understanding setting. In the unsupervised setting, we focus on reconstruction, prediction, and embedding-based models like MemAE~\cite{gong2019memorizing} and MNAD~\cite{park2020learning}, which use memory modules to enhance anomaly discrimination. For weakly-supervised anomaly detection, we adopt methods such as VadCLIP~\cite{vadclip} and MGFN~\cite{chen2023mgfn}, which use feature magnitudes and vision-language associations. In video-understanding, we evaluate video-language models like Video-ChatGPT~\cite{maaz2023video} and image-language models like CLIP~\cite{radford2021learning} to predict detailed anomaly descriptions and scores through cross-modal embeddings.

\noindent\textbf{Code and Experiment details.} We provide code and toolkit for our dataset and benchmark. More experiment details are listed in Appendix.

\begin{table}[t!]
\centering  \setlength{\abovecaptionskip}{0.1cm}
\caption{\textbf{Physics Anomaly Explanation (PAEval) metric results} on our Phys-AD dataset.}
\setlength{\tabcolsep}{3pt} 
\renewcommand\arraystretch{1.0}
\renewcommand{\arraystretch}{1.0} 
\resizebox{\linewidth}{!}{
\begin{tabular}{l|c|c|c}
\toprule
\textbf{Methods}   & \textbf{Classification} ($\uparrow$) & \textbf{Description} ($\uparrow$)& \textbf{Explanation} ($\uparrow$) \\ \midrule
LAVAD\cite{zanella2024harnessing}       &0.510     &0.157      &0.000            \\ 
Video-ChatGPT\cite{maaz2023video}&0.496     &0.131      &0.160            \\ 
Video-LLaMA\cite{zhang2023video} & \textbf{0.523}    &0.137      &\textbf{0.303}            \\ 
Video-LLaVA\cite{lin2023video} &0.463     &\textbf{0.219}      &0.282            \\     \bottomrule
\end{tabular}
}
\label{tab:5}\vspace{-2mm}
\end{table}

\subsection{Main Findings}

\noindent\textbf{Overall anomaly detection benchmarking results}. Table \ref{tab:4} shows that existing video anomaly detection and video-understanding methods achieve limited performance on the Phys-AD dataset, with the highest AUROC only reaching 66.9\% for MNAD.r~\cite{park2020learning}. This result underscores the heightened complexity of Phys-AD compared to existing datasets, highlighting a domain shift that challenges current industrial anomaly detection algorithms, which are often tuned to visually distinct anomaly patterns in single frames rather than complex temporal or physical cues.

\vspace{1mm}
\noindent\textbf{Anomaly detection via unsupervised and weakly supervised methods}. Our experiment includes unsupervised methods (e.g., MemAE~\cite{gong2019memorizing}, MPN~\cite{lv2021learning}, MNAD~\cite{park2020learning}) and weakly supervised methods (e.g., S3R~\cite{S3R}, MGFN~\cite{chen2023mgfn}, VADClip~\cite{vadclip}). Unsupervised methods like MNAD.p~\cite{park2020learning} performed better on temporal anomalies (81.0\% for fan, 68.0\% for screw) by leveraging prediction-based approaches, which excel in scenarios requiring temporal understanding. Weakly supervised methods improve baseline scores across complex classes, preventing extreme low scores in challenging categories such as spherical bearing. However, weakly supervised methods show marginally lower performance in simpler anomaly classes, indicating trade-offs introduced by anomaly samples in training.

\vspace{1mm}
\noindent\textbf{Unsupervised anomaly detection method performance by category}. Among unsupervised methods, MemAE~\cite{gong2019memorizing} achieves high AUROC on objects with spatial anomalies (e.g., sticky roller, 96.7\%) but underperforms in temporal anomaly classes like rolling and spherical bearings, where it achieves only 4.4\% and 9.2\%, respectively. MNAD~\cite{park2020learning} improves temporal sensitivity by incorporating temporal prototypes, achieving better scores for bearings but still struggling with complex anomaly types, revealing limitations in purely prototype-based approaches.

\vspace{1mm}
\noindent\textbf{Anomaly detection via Video-Language Models}. VLM-based methods (e.g.,Video-ChatGPT~\cite{maaz2023video}, Video-LLaMA~\cite{zhang2023video}) struggle, with top-performing Video-LLaMA~\cite{zhang2023video} reaching only 52.3\% average AUROC. Performance is impacted by reliance on pre-trained weights that are not optimized for physics-grounded video content, evident from low AUROC scores in categories with nuanced physical dynamics, such as hinges and screws. Further, PAEval results suggest that these models lack effective reasoning about object physical dynamics and behaviors influenced by physical forces, underscoring a gap between VLM capabilities and the demands of IAD tasks.

\vspace{1mm}
\noindent\textbf{Anomaly explanation via Video-Language Models.} Table \ref{tab:5} reports PAEval metric results, with the best-performing VLM achieving only 21.9\% in description and 30.3\% in explanation. These findings emphasize that current VLMs lack the depth in physical reasoning and temporal coherence required for understanding real-world physics-based scenarios, which our Phys-AD dataset demands.

\vspace{1mm}
\noindent\textbf{Summary}. Overall, the results reveal Phys-AD’s unique challenge in requiring both high spatial detail and temporal comprehension, areas where existing methods and models underperform. This analysis points to the need for future research in models that integrate temporal reasoning with physics-based anomaly detection.

\section{Conclusion}
In this paper, we introduce the first industrial anomaly detection task focusing on real-world scenarios where physical understanding and reasoning are essential for anomaly detection.  We present the Physics Anomaly Detection (Phys-AD) dataset, a large-scale, physics-grounded video dataset with over 6400 videos across 22 categories and 49 object types interacting with robotic systems, capturing 47 anomaly types that necessitate visual and physical understanding. We assess Phys-AD, highlighting the lack of baseline approaches for high-level reasoning in anomaly detection. Additionally, we propose the Physics Anomaly Explanation (PAEval) metric to evaluate visual language models (VLMs) on physics-based reasoning. Experiments show that current VLMs fall short of human-level understanding in physics-based anomaly scenarios. This work marks a milestone for industrial anomaly detection, promoting physics-grounded reasoning in complex industrial settings.

\vspace{1mm}
\noindent\textbf{Limitation and future work.}  Although our Phys-AD dataset provides a large variety of objects with diverse physical properties and various types of interaction methods, we plan to add even more diverse interaction methods and objects in the future to better meet the demands of complex real-world industrial scenarios. Due to the significant differences between our Phys-AD dataset and current industrial anomaly detection and video anomaly detection datasets, most existing anomaly detection algorithms cannot be directly applied to our dataset. In the future, we will test more algorithms on our Phys-AD dataset and provide experimental results across more settings like zero-shot, few-shot, semi-supervised settings, \textit{etc}.

\clearpage
{
    \small
    \bibliographystyle{ieeenat_fullname}

}



\clearpage
\setcounter{page}{1}
\maketitlesupplementary
\raggedbottom

\setcounter{table}{0}  
\setcounter{figure}{0}  
\setcounter{section}{0}  
\renewcommand{\thetable}{\Alph{table}}
\renewcommand{\thefigure}{\Alph{figure}}
\renewcommand{\thesection}{\Alph{section}}

In this supplementary material, we provide additional details on our benchmark methods, an in-depth benchmarking analysis, and further illustrations of the Phys-AD dataset with accompanying figures.

\section{More Benchmark Methods Details}

We provide comprehensive details on the benchmark methods used in our experiments. The methods are grouped into three categories: unsupervised anomaly detection, weakly supervised anomaly detection, and video-understanding-based methods.

\subsection{Unsupervised Anomaly Detection Methods}

We selected popular and reproducible video anomaly detection algorithms for the unsupervised setting, including reconstruction-based, prediction-based, and embedding-based methods.

\subsubsection{MemAE \cite{gong2019memorizing}}

MemAE is a pioneering work that introduces a memory module to an autoencoder for video frame reconstruction-based anomaly detection methods. It addresses the problem where autoencoders can sometimes reconstruct anomalous parts of the input.

\subsubsection{MPN \cite{lv2021learning}}

Based on MemAE, MPN proposes a Dynamic Prototype Unit (DPU) to encode normal dynamics as prototypes in real-time, eliminating extra memory costs.

\subsubsection{MNAD \cite{park2020learning}}

MNAD uses a memory module with a novel update scheme where items in the memory record prototypical patterns of normal data. It presents feature compactness and separateness losses to train the memory, enhancing the discriminative power of both memory items and learned features from normal data. We designed two experimental versions:

\begin{itemize}
    \item \textbf{MNAD.r}: Aimed at current frame reconstruction.
    \item \textbf{MNAD.p}: Aimed at future frame prediction.
\end{itemize}

\subsubsection{SVM \cite{sultani2018real}}

SVM extracts normal temporal features using a pre-trained I3D feature extractor and trains a Support Vector Machine to classify normal and abnormal features.

\subsection{Weakly Supervised AD Methods}

For the weakly supervised setting, we selected popular methods based on self-supervised learning, feature magnitudes, and clip-based approaches.

\subsubsection{S3R \cite{S3R}}

S3R models the feature distribution of normal and abnormal data by combining self-supervised learning with dictionary learning. The training features are used to train a binary classifier for anomaly detection.

\subsubsection{MGFN \cite{chen2023mgfn}}

MGFN proposes the Feature Amplification Mechanism to enhance the discriminativeness of feature magnitudes for anomaly detection.

\subsubsection{VAD-CLIP \cite{vadclip}}

VAD-CLIP leverages detailed associations between vision and language powered by CLIP and incorporates a dual-branch classifier for anomaly detection.

\subsection{Video Understanding Methods}

We also evaluated video-language models and image-language models for video understanding and anomaly detection.

\subsubsection{Video-LLaMA \cite{zhang2023video}, Video-ChatGPT \cite{maaz2023video}, Video-LLaVA \cite{lin2023video}}

For these models, we directly fed designed prompts and videos to obtain descriptions and detection results.

\subsubsection{LAVAD \cite{zanella2024harnessing}}

LAVAD leverages CLIP to describe each frame's content and aggregates the descriptions of multiple frames to get the final video description.

\subsubsection{Zero-Shot CLIP \cite{radford2021learning} and Zero-Shot ImageBind \cite{girdhar2023imagebind}}

These methods utilize CLIP and ImageBind to project text features and image features into the same space, directly comparing the image features extracted by the pre-trained models with normal text features to obtain anomaly scores.

\section{More Visualization}

We provide further illustrations of the Phys-AD dataset through various figures, highlighting the unique characteristics and challenges it presents.

\subsection{Phys-AD Dataset Overview}

Figure~\ref{fig:3} shows examples from the Phys-AD dataset. Many deformable, articulated, and assembled objects appear normal in a static state but reveal anomalies only through physical manipulation.

\begin{figure*}[t!]
  \centering
  \includegraphics[width=1.0\textwidth]{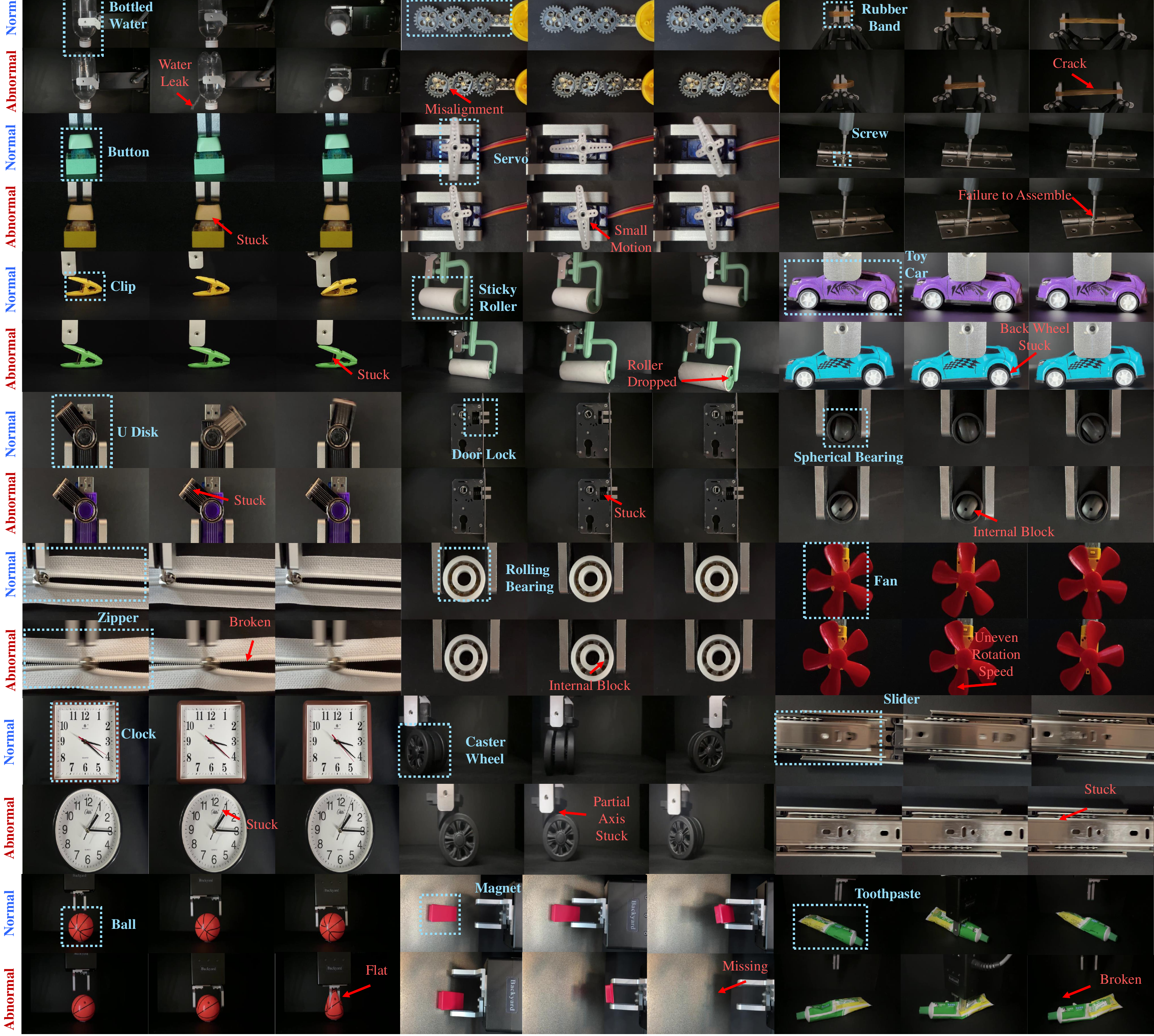}
  \caption{\textbf{Phys-AD Dataset.} Many deformable, articulated, and assembled objects appear normal in a static state but reveal anomalies only through physical manipulation.}
  \label{fig:3}
\end{figure*}

\subsection{Anomaly Cases}

\begin{figure*}[t!]
  \centering
  \setlength{\abovecaptionskip}{0.1cm}
  \includegraphics[width=0.95\textwidth, keepaspectratio]{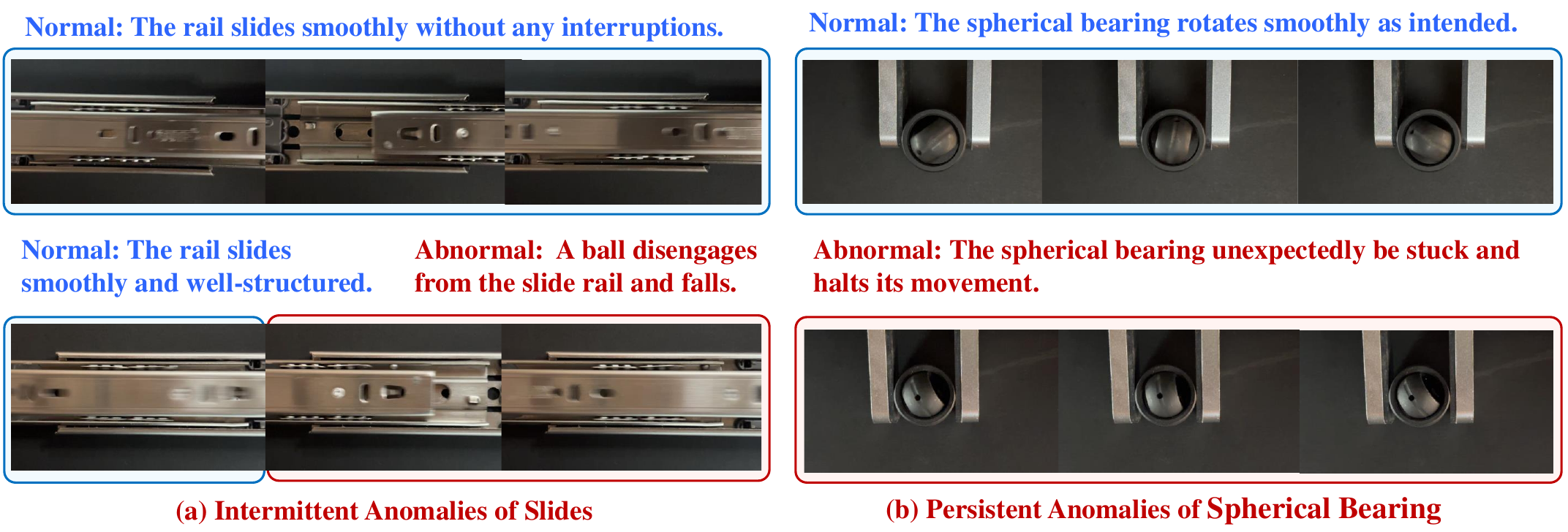}
  \includegraphics[width=0.95\textwidth, keepaspectratio]{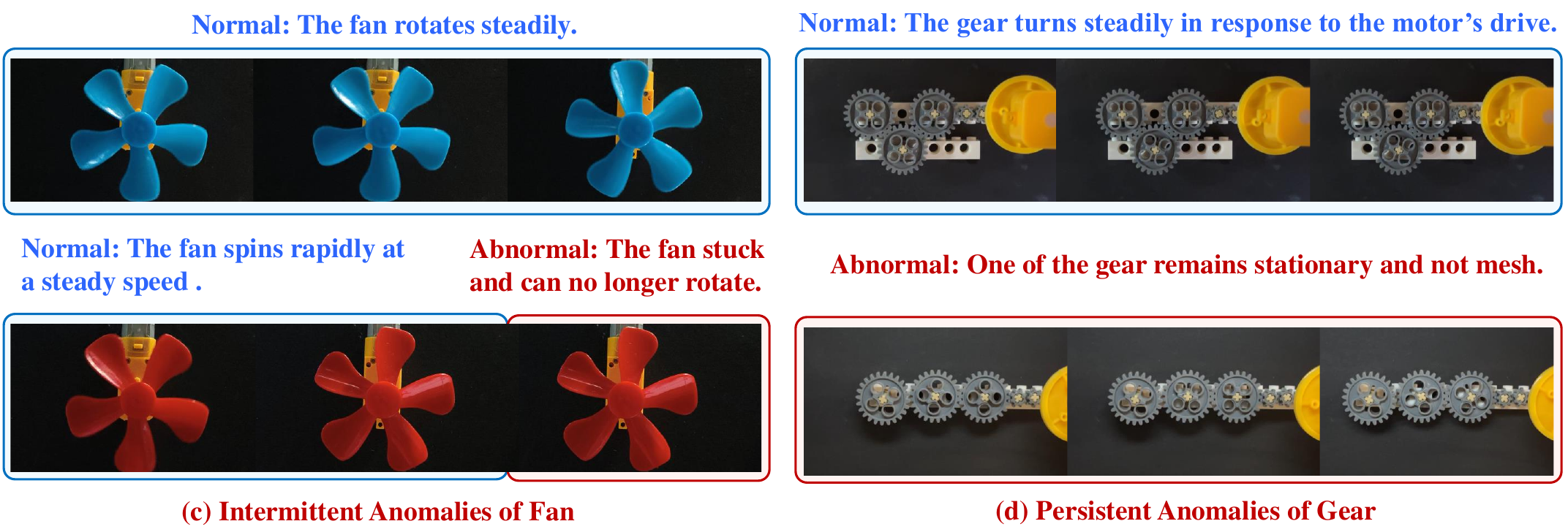}
  \includegraphics[width=0.95\textwidth, keepaspectratio]{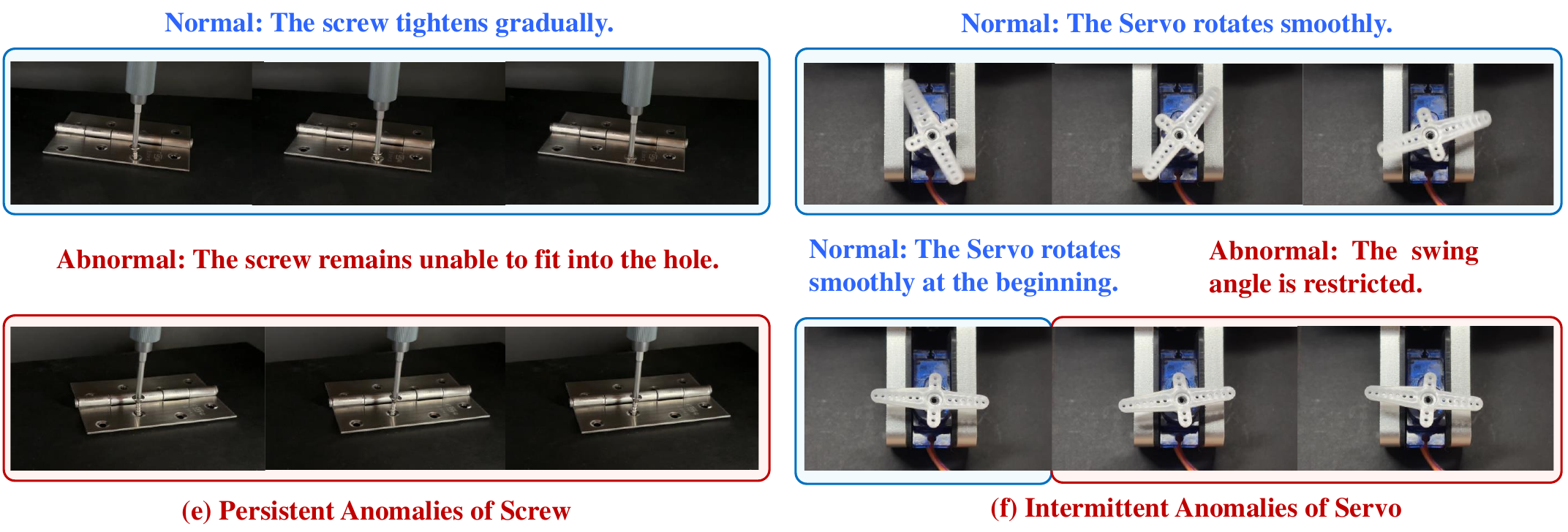}
  \includegraphics[width=0.95\textwidth, keepaspectratio]{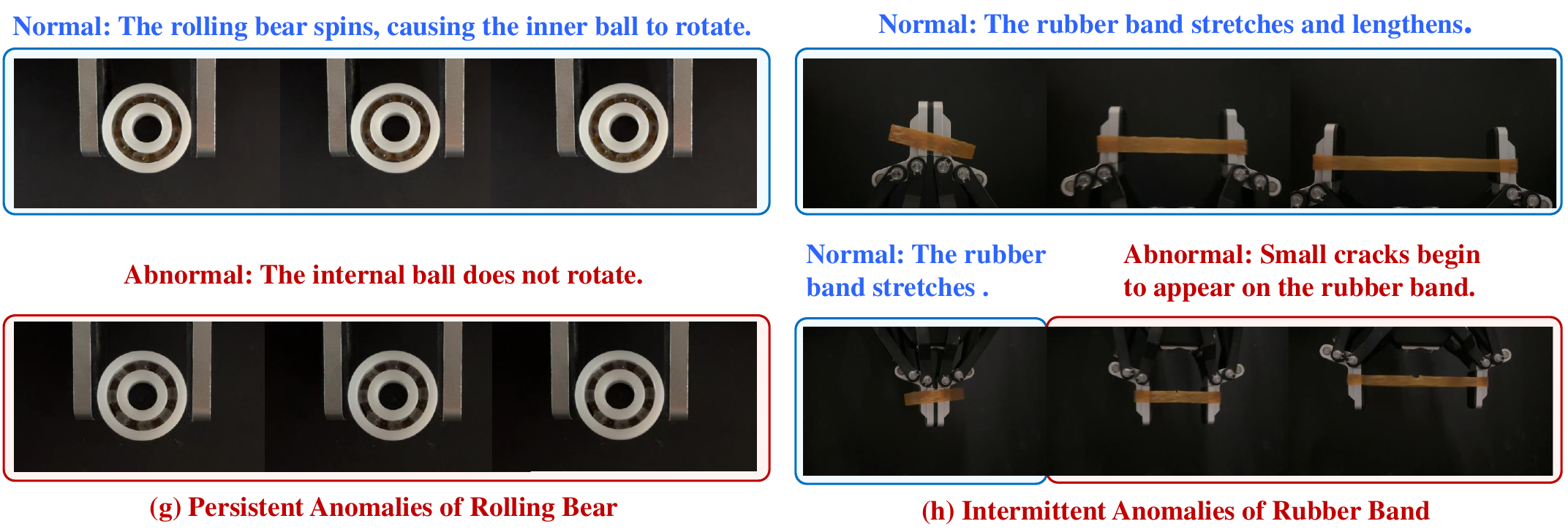}
  \caption{\textbf{Anomaly Cases of Phys-AD Dataset (1/2).}}
  \label{fig:8}
\end{figure*}

\begin{figure*}[t!]
  \centering
  \setlength{\abovecaptionskip}{0.1cm}
  \includegraphics[width=0.95\textwidth, keepaspectratio]{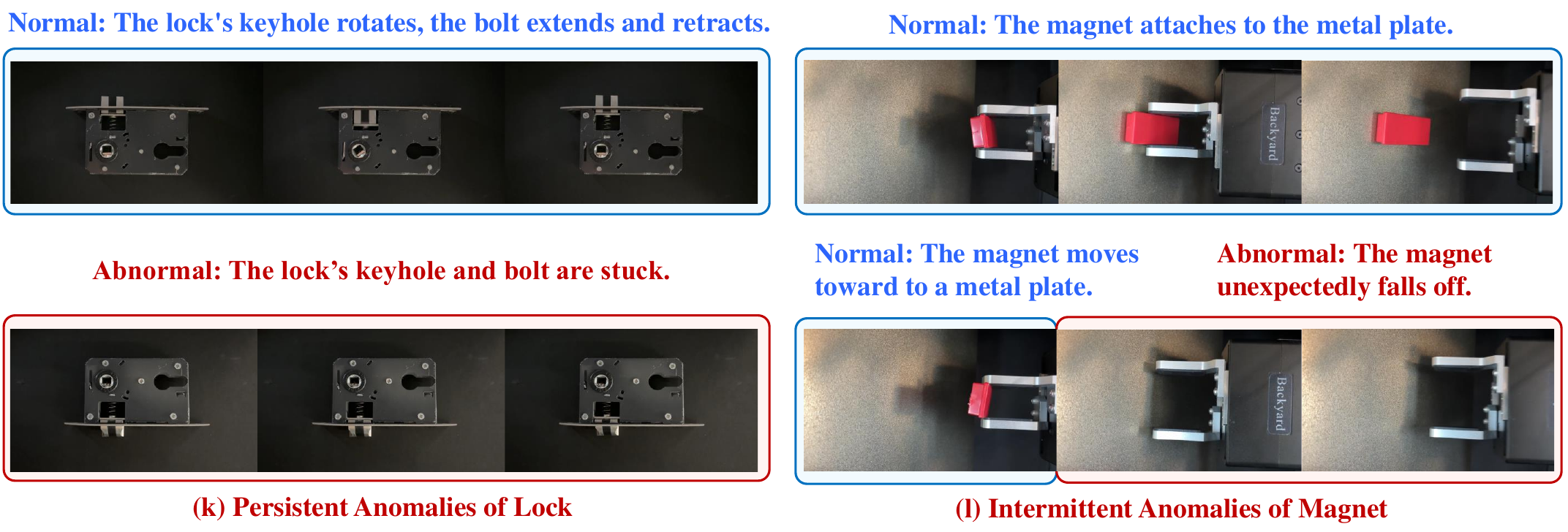}
  \includegraphics[width=0.95\textwidth, keepaspectratio]{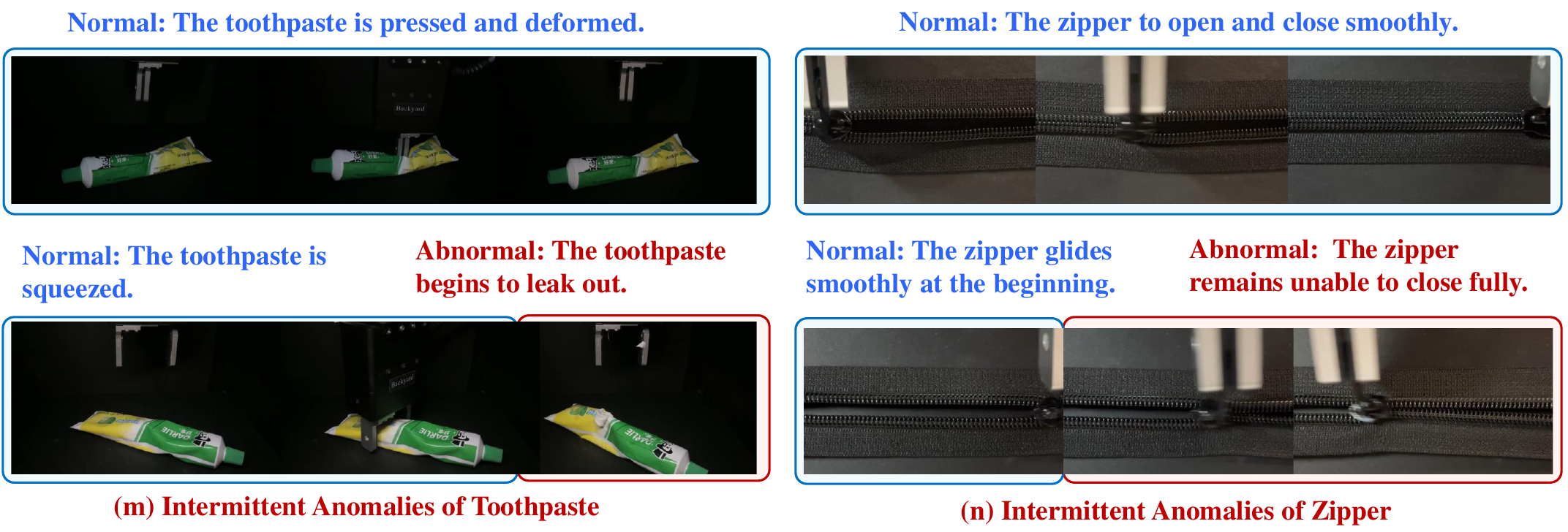}
  \includegraphics[width=0.95\textwidth, keepaspectratio]{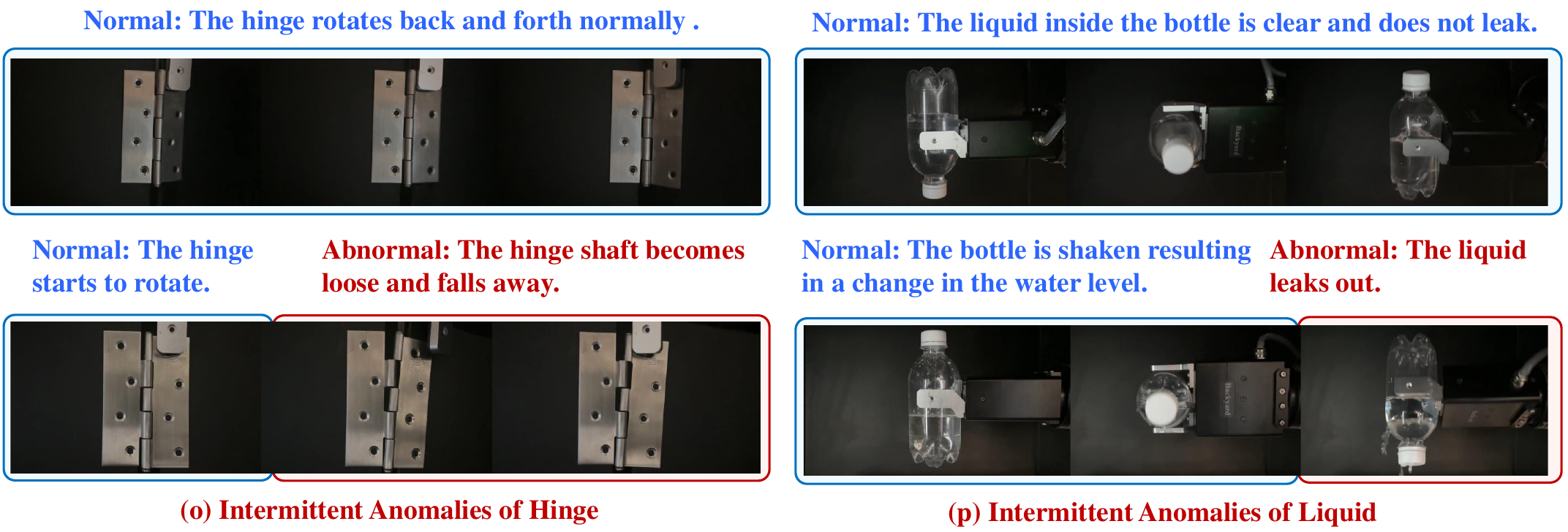}
  \includegraphics[width=0.95\textwidth, keepaspectratio]{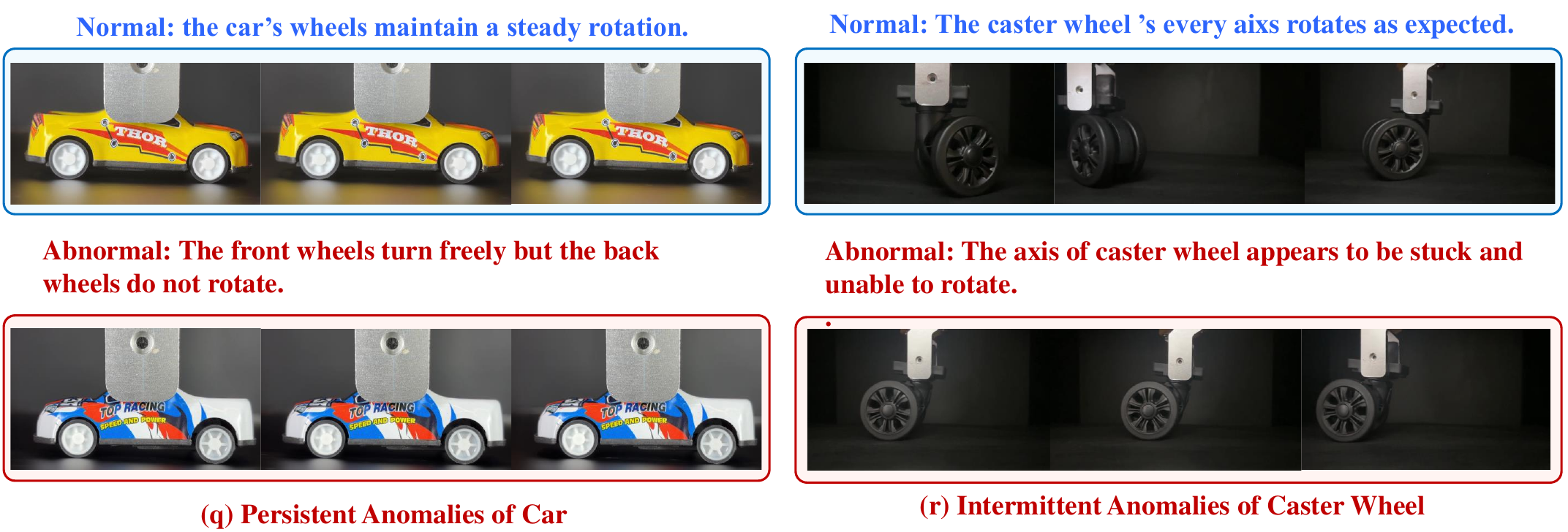}
  \caption{\textbf{Anomaly Cases of Phys-AD Dataset (2/2).}}
  \label{fig:9}
\end{figure*}

Figures~\ref{fig:8} and \ref{fig:9} present anomaly cases from the Phys-AD dataset. These figures illustrate various anomalies that are challenging to detect due to their subtle visual cues and reliance on physical interactions.

\subsection{Video Demonstrations}

\begin{figure*}[t]
  \centering
  \setlength{\abovecaptionskip}{0.1cm}

  \begin{subfigure}[b]{1.0\textwidth}
    \centering
    \includegraphics[width=\textwidth]{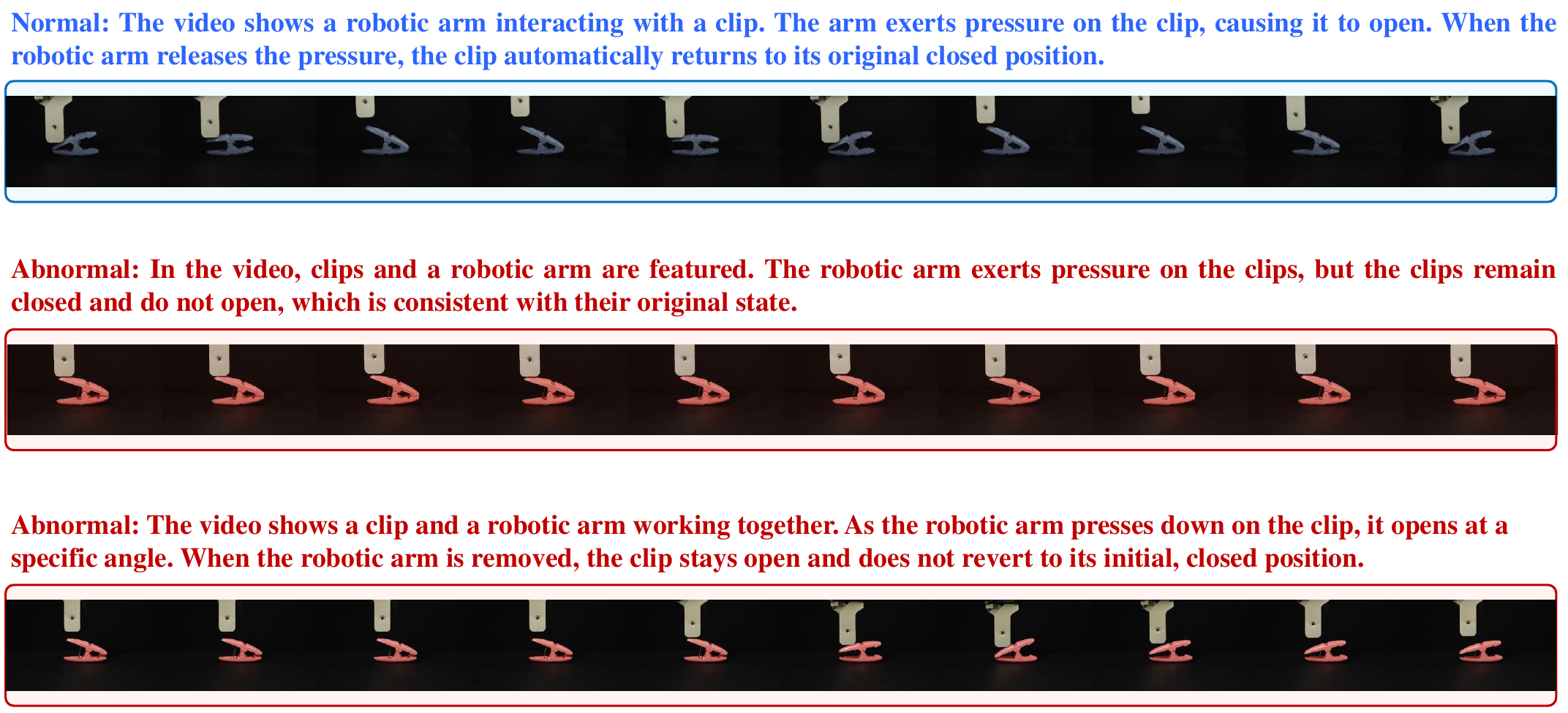}
    \caption{Video Demo of Clip}
    \label{fig:10a}
  \end{subfigure}
  \kern-1em
  \begin{subfigure}[b]{1.0\textwidth}
    \centering
    \includegraphics[width=\textwidth]{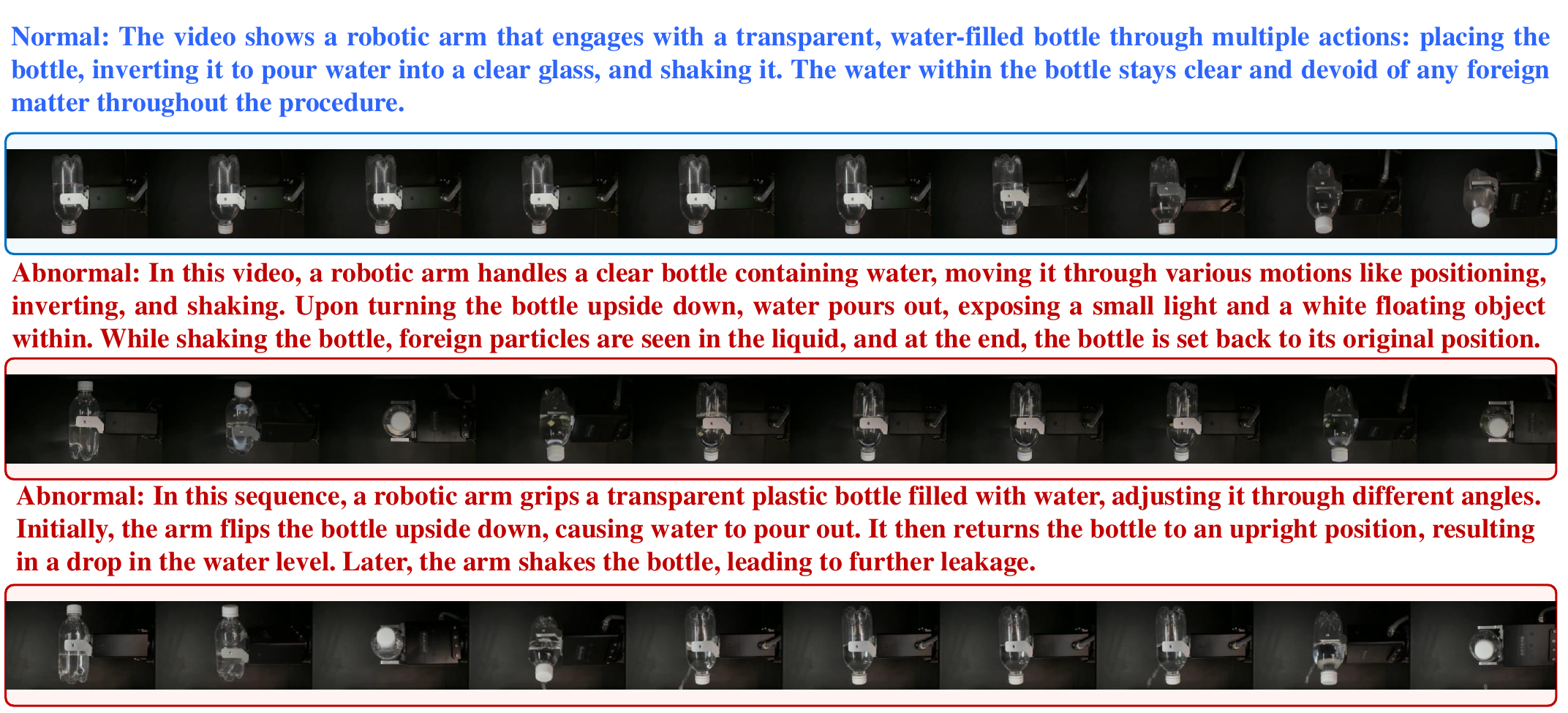}
    \caption{Video Demo of Liquid}
    \label{fig:10b}
  \end{subfigure}
  \kern-1em
  \begin{subfigure}[b]{1.0\textwidth}
    \centering
    \includegraphics[width=\textwidth, trim={0cm 2.5cm 0cm 2.5cm}, clip]{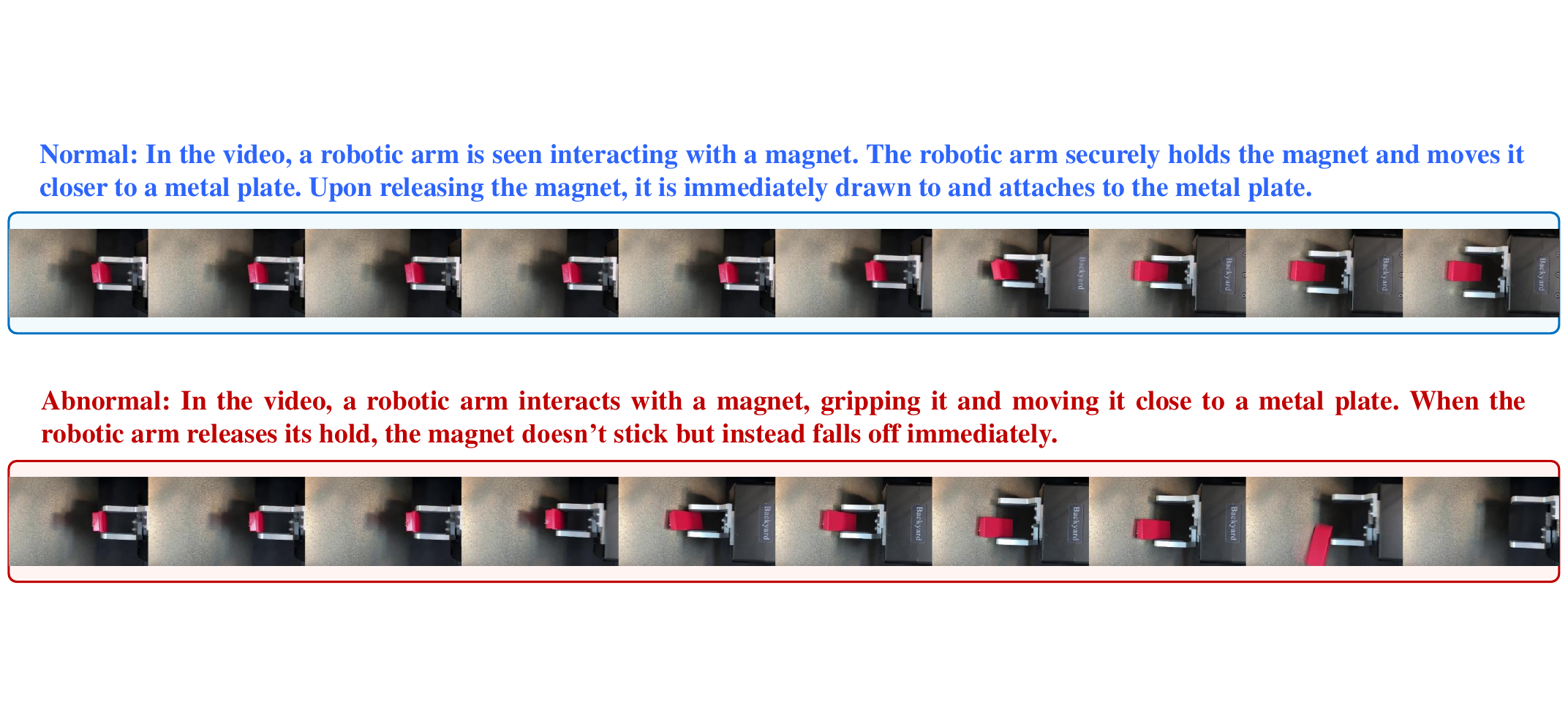}
    \caption{Video Demo of Magnet}
    \label{fig:10c}
  \end{subfigure}
  \caption{\textbf{Video Demo of the Phys-AD Dataset (1/2).}}
  \label{fig:10}
\end{figure*}

\begin{figure*}[t!]
  \centering
  \setlength{\abovecaptionskip}{0.1cm}
  \begin{subfigure}[b]{1.0\textwidth}
    \centering
    \includegraphics[width=\textwidth, trim={0cm 2.5cm 0cm 2.5cm}, clip]{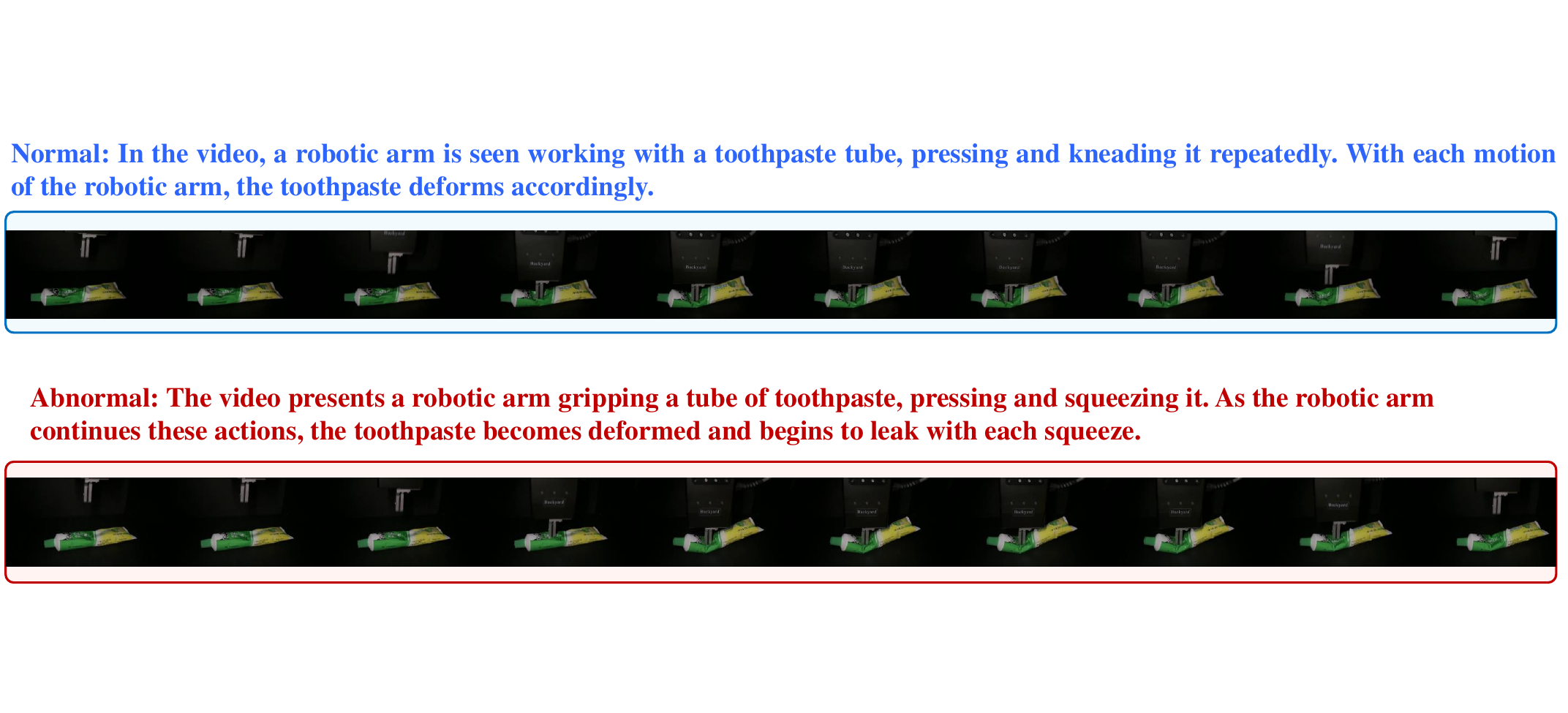}
    \caption{Video Demo of Toothpaste}
    \label{fig:11a}
  \end{subfigure}
  \begin{subfigure}[b]{1.0\textwidth}
    \centering
    \includegraphics[width=\textwidth, trim={0cm 2.5cm 0cm 2.5cm}, clip]{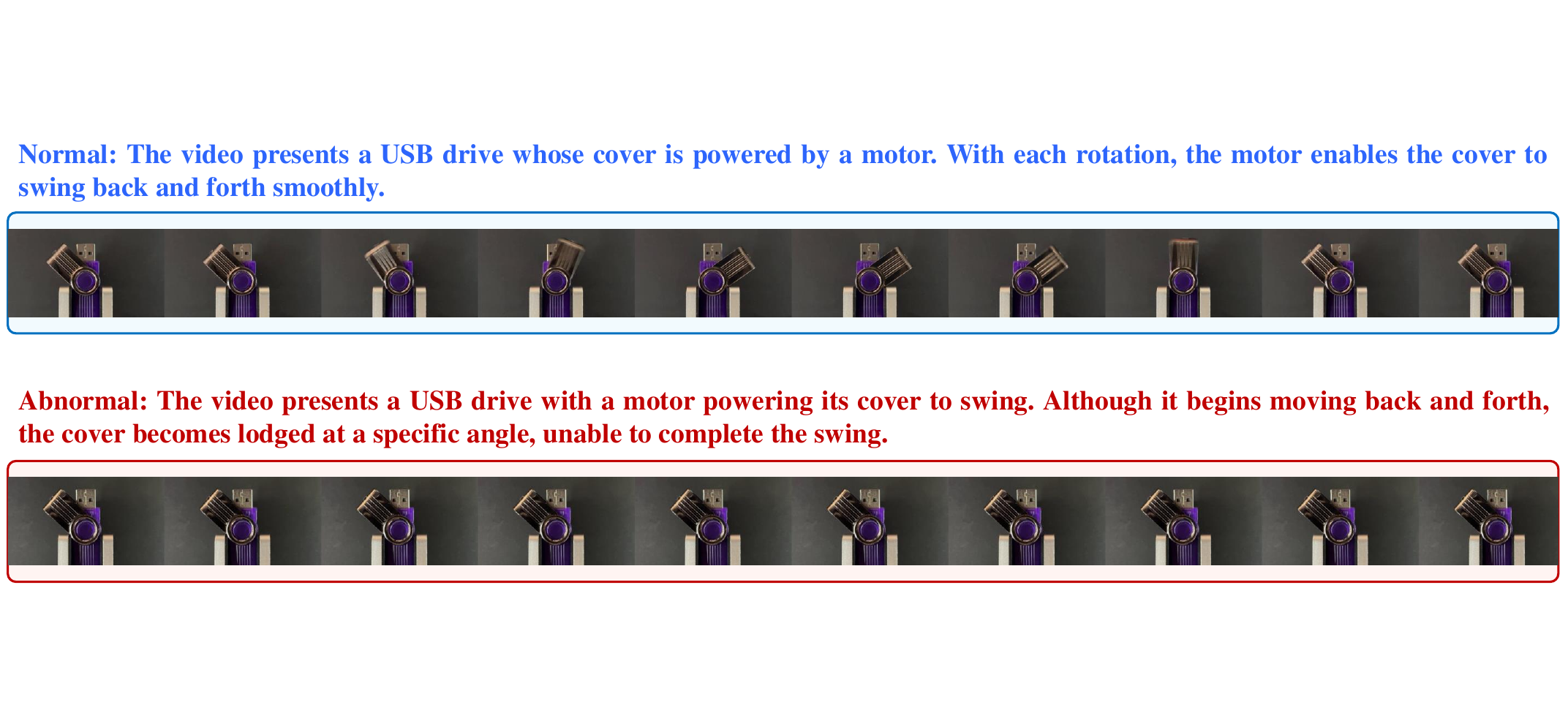}
    \caption{Video Demo of USB}
    \label{fig:11b}
  \end{subfigure}
  \caption{\textbf{Video Demo of the Phys-AD Dataset (2/2).}}
  \label{fig:11}
\end{figure*}

Figures~\ref{fig:10} and \ref{fig:11} provide video demonstrations from the Phys-AD dataset, showcasing dynamic interactions that highlight the physical properties and anomalies present in the data.

\subsection{Generation of PAEval Labels}

\begin{figure*}[h]
  \centering
  \setlength{\abovecaptionskip}{0.1cm}
  \includegraphics[width=0.95\textwidth]{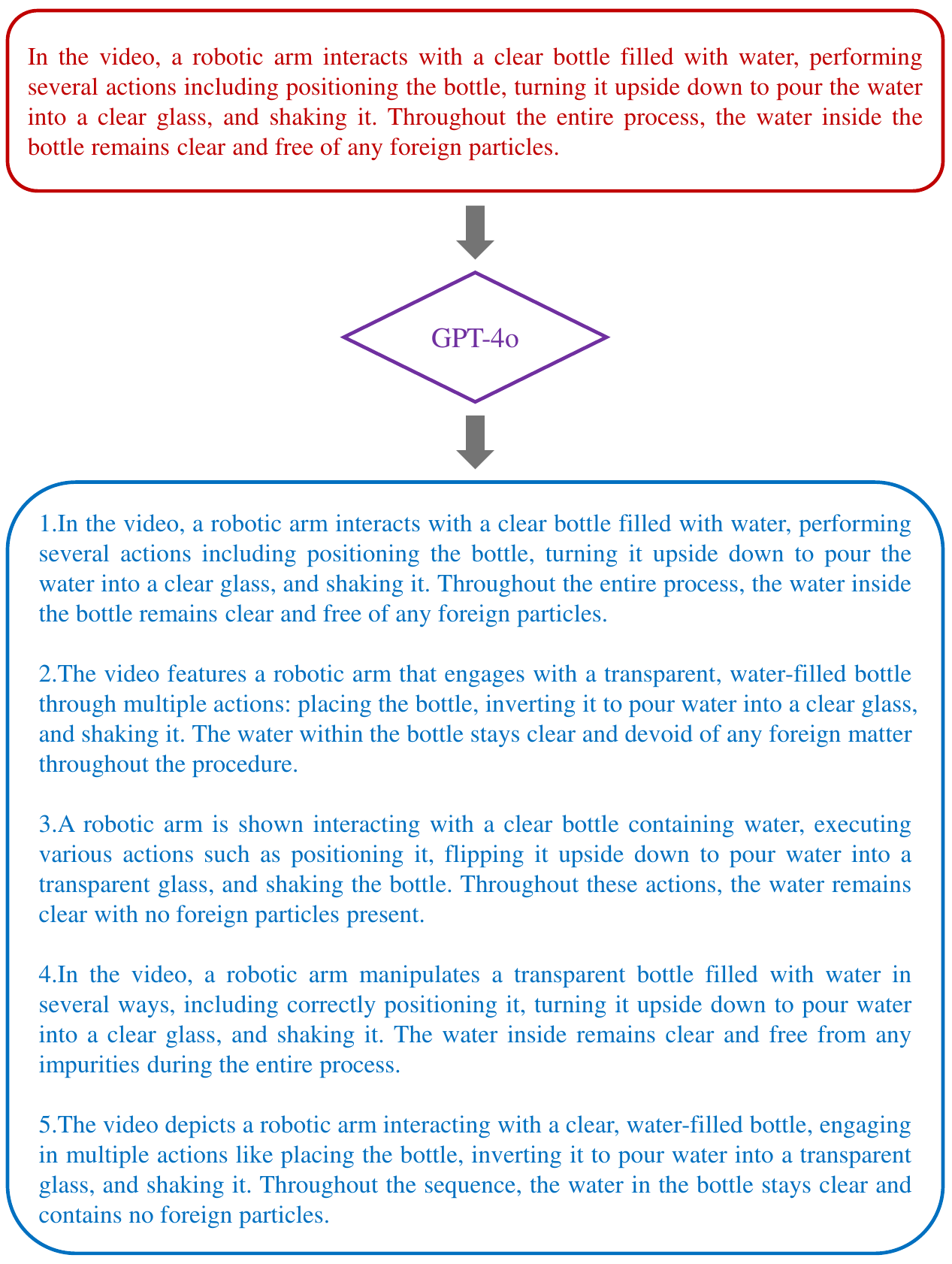}
  \caption{\textbf{Generation of PAEval Labels (Labels for Liquid in the Graph).}}
  \label{fig:12}
\end{figure*}

Figure~\ref{fig:12} illustrates the generation of PAEval labels, demonstrating how labels for physical properties are generated in the dataset.

\subsection{Examples of Video-Language Model Descriptions}

\begin{figure*}[h]
  \centering
  \setlength{\abovecaptionskip}{0.1cm}
  \includegraphics[width=1.0\textwidth]{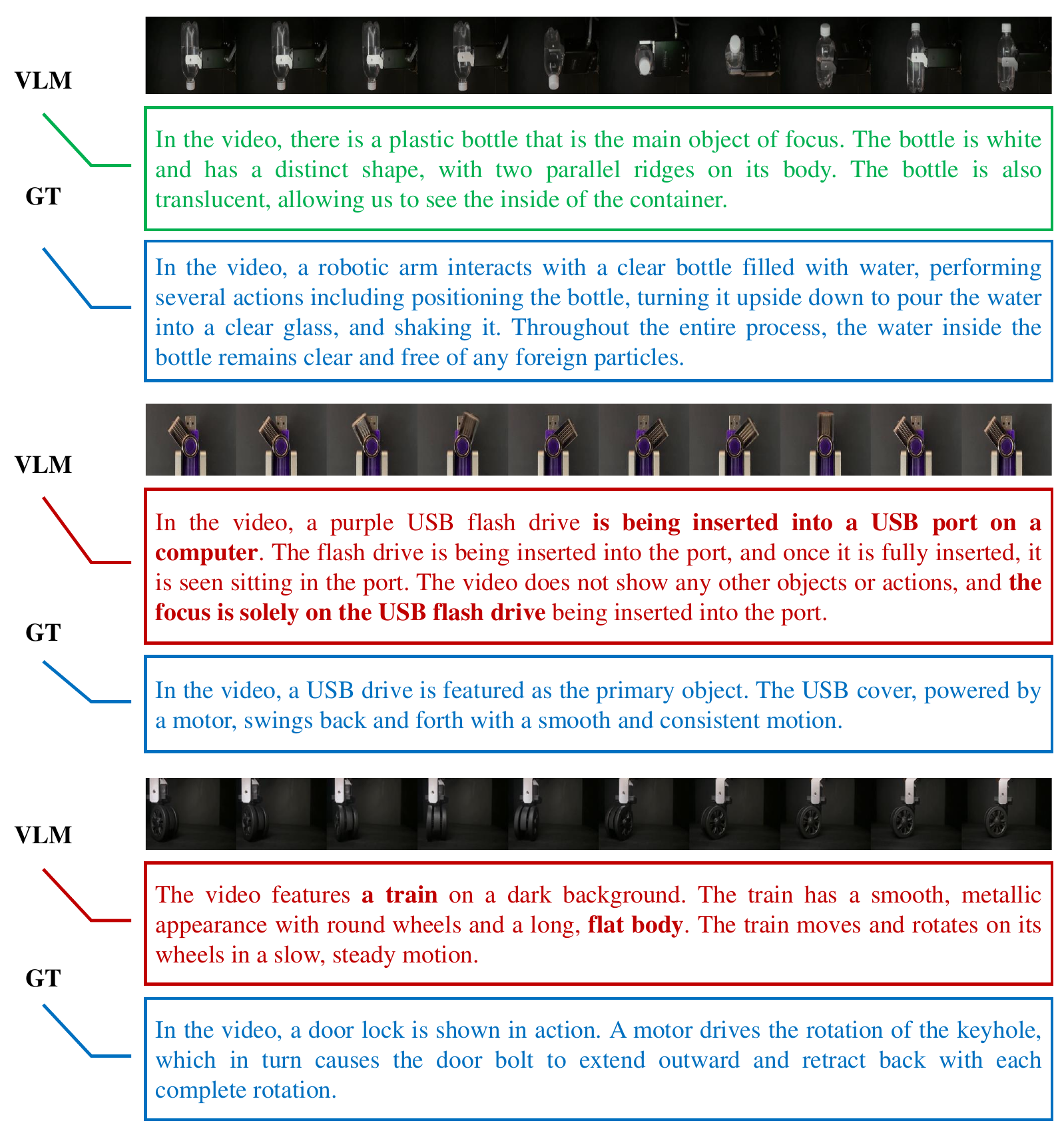}
 \vspace{2mm} \caption{\textbf{Examples of anomaly descriptions generated by Video-LLaMA.} {Green text} indicates correct summaries, while {red text} indicates incorrect ones.}

  \label{fig:13}
\end{figure*}

Figure~\ref{fig:13} shows examples of how video-language models describe videos from the Phys-AD dataset. Correct summaries are highlighted in green, and incorrect ones in red, emphasizing the challenges VLMs face in understanding these videos.

\section{Implementation Details}
\subsection{Hyperparameters}
The modified hyperparameters are detailed below, while others remain the same as specified in the original works:
\begin{enumerate}
    \item \textbf{MemAE}~\cite{gong2019memorizing}: Batch size = 32, epochs = 30
    \item \textbf{MPN}~\cite{lv2021learning}: Epochs = 21
    \item \textbf{MNAD}~\cite{park2020learning}: Epochs = 7
    \item \textbf{S3R}~\cite{S3R}: Batch size = 4, epochs = 1000
    \item \textbf{MGFN}~\cite{chen2023mgfn}: Epochs = 500
\end{enumerate}

\subsection{Prompts}
We employed various prompt designs to guide models, as described below.

\subsubsection{\textbf{LAVAD}~\cite{zanella2024harnessing}}
\begin{itemize}
    \item \textbf{Score Prompt:} "If you were a law enforcement agency, how would you rate the scene described on a scale from 0 to 1, with 0 representing a standard scene and 1 denoting a scene with suspicious activities? Provide the response as a Python list, selecting a single number from \texttt{[0, 0.1, 0.2, ..., 1.0]}. Do not include any textual explanation."
    \item \textbf{Description Prompt:} "Summarize the events described in the temporal sequence of the scene. Avoid unnecessary details or elaboration."
\end{itemize}

\subsubsection{\textbf{Video-LLaVA}~\cite{lin2023video}}
\begin{itemize}
    \item \textbf{Score Prompt:} "Evaluate the anomaly level in this video on a scale from 0 to 1, where 0 represents 'Absolutely no abnormalities' and 1 represents 'Absolutely abnormal.' Avoid using the exact score of 0.5. Provide only the score without additional explanation."
    \item \textbf{Description Prompt:} "Describe the events depicted in the video in detail."
    \item \textbf{Explanation Prompt:} "Assume you are an anomaly detection expert. Is the function of the object in the video normal or abnormal? Provide a detailed explanation."
\end{itemize}

\subsubsection{\textbf{Video-LLaMA}~\cite{zhang2023video} \& \textbf{Video-ChatGPT}~\cite{maaz2023video}}
For these models, prompts consist of \texttt{COMMON\_QUESTIONS}, followed by \texttt{CATEGORY\_SPECIFIC\_QUESTIONS}, depending on the object class. Scoring, description, and explanation are generated in one step. Below is an example for the \textit{ball} class:
\begin{itemize}
    \item \texttt{COMMON\_QUESTIONS:}
    \begin{enumerate}
        \item "What is the object in the video?"
        \item "What is the normal function of the object in real life?"
        \item "What is the mode of interaction observed in the video?"
        \item "Describe the content of this video, focusing on objects, appearance, and physical interactions."
        \item "As an anomaly detection expert, assess whether the object's function is normal or abnormal. Provide a reasonable explanation."
    \end{enumerate}
    \item \texttt{CATEGORY\_SPECIFIC\_QUESTIONS (for ball class):} "Assume the object in the video is a ball. Under normal conditions, a fully inflated ball resists significant deformation. Rate the anomaly on a scale from 0 to 1, where 0 is 'definitely normal' and 1 is 'definitely abnormal.' Provide only the anomaly score in the format: \texttt{\{anomalyscore=\}} without additional text."
\end{itemize}

\subsubsection{\textbf{PAEval Prompt}}
\begin{itemize}
    \item \textbf{System Prompt:} "I am an expert in text comparison. I evaluate the semantic similarity of texts, considering spatiotemporal relationships and event structures. I assign a similarity score between 0 and 1, where higher scores indicate greater similarity."
    \item \textbf{User Prompt:} "Given the input text: \{\texttt{text}\}, compare it to entries in the label text library: \{\texttt{labels}\}. Assign a similarity score and output only the highest score as the result."
\end{itemize}

\section{Additional Experimental Results}

\subsection{Result Overview}
Tables~\ref{tab:A} and \ref{tab:B} summarize the Average Precision (AP) and Accuracy (ACC) of various methods on the Phys-AD dataset across 22 categories. These results cover three methodological paradigms: unsupervised, weakly supervised, and video understanding approaches. It is important to note that both AP and ACC metrics can be influenced by the ratio of positive and negative samples, making these metrics indicative rather than absolute. For ACC, a decision threshold of 0.5 is used by default.

\subsection{Observations and Insights}

\subsubsection{Performance Trends Across Paradigms}
\begin{itemize}
    \item \textbf{Unsupervised Methods:} These approaches, such as MNAD.r, achieve competitive results in simpler scenarios, with an average AP of 0.797. However, they often struggle with categories that exhibit complex temporal dynamics or physical interactions.
    \item \textbf{Weakly Supervised Methods:} Methods like S3R and MGFN outperform unsupervised approaches, benefiting from limited supervision. They show consistent improvements in categories requiring higher precision.
    \item \textbf{Video Understanding Models:} Advanced models such as Video-LLaMA and Video-LLaVA demonstrate superior performance by leveraging contextual and semantic reasoning. This is evident in challenging categories such as `Car' and `Gear,' where contextual understanding plays a key role.
\end{itemize}

\subsubsection{Category-Specific Insights}
\begin{itemize}
    \item \textbf{High Variability:} Categories like `Sticky Roller' and `Servo' exhibit significant performance variability across methods, highlighting the challenges of modeling subtle interactions and anomalies.
    \item \textbf{Limitations in Specific Categories:} Categories like `Rubber Band' and `USB' present low AP and ACC across all methods, reflecting the difficulty of detecting low-contrast anomalies or deformations.
    \item \textbf{Strengths in Contextual Modeling:} In categories like `Ball' and `Magnet,' video understanding models excel, showcasing the advantage of integrating physical reasoning and contextual cues.
\end{itemize}

\subsubsection{Challenges with Balanced Metrics}
\begin{itemize}
    \item The ACC metric is highly sensitive to class imbalance, particularly in categories like `Hinge' and `Caster Wheel,' where unsupervised methods often underperform due to skewed distributions.
\end{itemize}

\subsubsection{General Observations}
\begin{itemize}
    \item \textbf{Incorporation of Domain Knowledge:} Models incorporating domain-specific knowledge, such as Video-LLaMA and LAVAD, perform significantly better in categories like `Button' and `Clip.'
    \item \textbf{Plateaus in Weakly Supervised Performance:} While effective, weakly supervised methods may reach performance ceilings, suggesting the need for more advanced hybrid or fully supervised approaches.
\end{itemize}

\section{Potential Negative Social Impacts}

Our dataset was collected with permission from the factory, ensuring compliance with ethical standards. Therefore, we anticipate no negative social impacts arising from this work.

\begin{table*}[t!]
\centering
\caption{\textcolor{black}{\textbf{Video-level AP ($\uparrow$) result of 22 categories on Phys-AD dataset.} We include Unsupervised, Weakly-supervised and Video-understanding methods.‘ZS ImgB’,‘V-ChatGPT',‘V-LLaMA',‘V-LLaVA' denote ZS ImageBind,Video-ChatGPT,Video-LLaMA and Video-LLaVA.}}
\vspace{-5pt}
\centering\setlength{\tabcolsep}{1mm}
\resizebox{\textwidth}{!}{
\begin{tabular}{l|ccccc|ccc|cccccc}
\toprule
\multirow{2}{*}{\textbf{Category.}} & 
    \multicolumn{5}{c|}{\textbf{Unsupervised}} & 
    \multicolumn{3}{c|}{\textbf{Weakly-supervised}} &
    \multicolumn{6}{c}{\textbf{Video-understanding}} \\
    \cmidrule(l){2-6} 
    \cmidrule(l){7-9} 
    \cmidrule(l){10-15}
& \makecell[c]{MPN\\\cite{lv2021learning}} & \makecell[c]{MemAE\\\cite{gong2019memorizing}} & \makecell[c]{MNAD.p\\\cite{park2020learning}} & \makecell[c]{MNAD.r\\\cite{park2020learning}} & \makecell[c]{SVM\\\cite{sultani2018real}} & \makecell[c]{VADClip\\\cite{vadclip}} & \makecell[c]{S3R\\\cite{S3R}} & \makecell[c]{MGFN\\\cite{chen2023mgfn}} & \makecell[c]{LAVAD\\\cite{zanella2024harnessing}} & \makecell[c]{ZS Clip\\\cite{radford2021learning}} & \makecell[c]{ZS ImgB\\\cite{girdhar2023imagebind}} & \makecell[c]{V-ChatGPT\\\cite{maaz2023video}} & \makecell[c]{V-LLaMA\\\cite{zhang2023video}} & \makecell[c]{V-LLaVA\\\cite{lin2023video}} \\ \midrule

Car               &0.628  &0.770  &0.762  &0.981  &0.784    &0.787  &0.816  &0.797       &0.773 &0.750 &0.750 &0.751 &0.876&0.759  \\ 
Fan               &0.916  &0.371  &0.933  &0.763  &0.750    &0.800  &0.823  &0.796       &0.757 &0.750 &0.750 &0.763 &0.861&0.795  \\   
Rolling Bearing   &0.522  &0.320  &0.479  &0.812  &0.882    &0.551  &0.619  &0.648       &0.499 &0.500 &0.500 &0.429 &0.741&0.500  \\ 
Spherical Bearing &0.332  &0.382  &0.929  &0.787  &0.588    &0.500  &0.647  &0.596       &0.488 &0.500 &0.500 &0.479 &0.693&0.500  \\ 
Servo             &0.745  &0.795  &0.992  &0.957  &0.750    &0.759  &0.803  &0.808       &0.756 &0.750 &0.750 &0.753 &0.857&0.745  \\  
Clip              &0.732  &0.619  &0.785  &0.596  &0.667    &0.631  &0.722  &0.720       &0.701 &0.693 &0.667 &0.761 &0.820&0.649  \\  
USB               &0.357  &0.742  &0.646  &0.946  &0.500    &0.530  &0.567  &0.548       &0.512 &0.500 &0.500 &0.535 &0.765&0.500  \\   
Hinge             &0.924  &0.965  &0.887  &0.967  &0.750    &0.868  &0.789  &0.834       &0.794 &0.758 &0.750 &0.750 &0.894&0.750  \\   
Sticky Roller     &0.989  &0.989  &0.698  &0.975  &0.667    &0.686  &0.883  &0.829       &0.553 &0.625 &0.667 &0.645 &0.834&0.656  \\   
Caster Wheel      &0.730  &0.815  &0.644  &0.823  &0.750    &0.797  &0.876  &0.901       &0.820 &0.735 &0.750 &0.732 &0.891&0.750  \\   
Screw             &0.522  &0.763  &0.750  &0.685  &0.667    &0.667  &0.769  &0.720       &0.763 &0.680 &0.667 &0.658 &0.826&0.690  \\   
Lock              &0.741  &0.683  &0.623  &0.795  &0.789    &0.662  &0.704  &0.767       &0.586 &0.667 &0.667 &0.613 &0.831&0.667  \\   
Gear              &0.839  &0.826  &0.874  &0.865  &0.800    &0.807  &0.829  &0.818       &0.603 &0.800 &0.800 &0.816 &0.903&0.805  \\   
Clock             &0.572  &0.751  &0.614  &0.711  &0.670    &0.670  &0.708  &0.698       &0.684 &0.667 &0.670 &0.667 &0.842&0.670  \\   
Slide             &0.806  &0.991  &0.978  &0.942  &0.667    &0.817  &0.844  &0.864       &0.772 &0.800 &0.800 &0.817 &0.906&0.726  \\  
Zipper            &0.898  &0.669  &0.896  &0.674  &0.667    &0.669  &0.777  &0.757       &0.712 &0.667 &0.667 &0.715 &0.832&0.667  \\   
Button            &0.966  &0.726  &0.903  &0.853  &0.800    &0.845  &0.818  &0.830       &0.778 &0.842 &0.800 &0.804 &0.905&0.763  \\   
Liquid            &0.564  &0.890  &0.860  &0.927  &0.667    &0.686  &0.726  &0.900       &0.712 &0.784 &0.667 &0.622 &0.800&0.579  \\  
Rubber Band       &0.411  &0.410  &0.433  &0.394  &0.536    &0.491  &0.670  &0.631       &0.499 &0.500 &0.500 &0.509 &0.751&0.478  \\  
Ball              &0.716  &0.661  &0.842  &0.826  &0.667    &0.667  &0.809  &0.727       &0.739 &0.667 &0.667 &0.699 &0.839&0.682  \\  
Magnet            &0.746  &0.754  &0.802  &0.603  &0.667    &0.667  &0.737  &0.841       &0.673 &0.667 &0.667 &0.780 &0.860&0.626  \\   
Toothpaste        &0.657  &0.899  &0.653  &0.644  &0.500    &0.500  &0.682  &0.746       &0.569 &0.500 &0.500 &0.464 &0.763&0.484  \\  \midrule
Average           &0.703  &0.735  &0.772  &0.797  &0.690    &0.684  &0.755  &0.763       &0.681 &0.673 &0.666 &0.671 &0.831&0.656  \\  \bottomrule
\end{tabular}
}
\label{tab:A}\vspace{-2mm}
\end{table*}

\begin{table*}[t!]
\centering
\caption{\textcolor{black}{\textbf{Video-level ACC ($\uparrow$) result of 22 categories on Phys-AD dataset.} We include Unsupervised, Weakly-supervised and Video-understanding methods.‘ZS ImgB’,‘V-ChatGPT',‘V-LLaMA',‘V-LLaVA' denote ZS ImageBind,Video-ChatGPT,Video-LLaMA and Video-LLaVA.}}
\vspace{-5pt}
\centering\setlength{\tabcolsep}{1mm}
\resizebox{\textwidth}{!}{
\begin{tabular}{l|ccccc|ccc|cccccc}
\toprule
\multirow{2}{*}{\textbf{Category.}} & 
    \multicolumn{5}{c|}{\textbf{Unsupervised}} & 
    \multicolumn{3}{c|}{\textbf{Weakly-supervised}} &
    \multicolumn{6}{c}{\textbf{Video-understanding}} \\
    \cmidrule(l){2-6} 
    \cmidrule(l){7-9} 
    \cmidrule(l){10-15}
& \makecell[c]{MPN\\\cite{lv2021learning}} & \makecell[c]{MemAE\\\cite{gong2019memorizing}} & \makecell[c]{MNAD.p\\\cite{park2020learning}} & \makecell[c]{MNAD.r\\\cite{park2020learning}} & \makecell[c]{SVM\\\cite{sultani2018real}} & \makecell[c]{VADClip\\\cite{vadclip}} & \makecell[c]{S3R\\\cite{S3R}} & \makecell[c]{MGFN\\\cite{chen2023mgfn}} & \makecell[c]{LAVAD\\\cite{zanella2024harnessing}} & \makecell[c]{ZS Clip\\\cite{radford2021learning}} & \makecell[c]{ZS ImgB\\\cite{girdhar2023imagebind}} & \makecell[c]{V-ChatGPT\\\cite{maaz2023video}} & \makecell[c]{V-LLaMA\\\cite{zhang2023video}} & \makecell[c]{V-LLaVA\\\cite{lin2023video}} \\ \midrule

Car               &0.703  &0.750  &0.755  &0.753  &0.793    &0.402  &0.262  &0.250       &0.332 &0.750 &0.750 &0.687 &0.504&0.447  \\ 
Fan               &0.825  &0.750  &0.769  &0.750  &0.750    &0.785  &0.302  &0.250       &0.311 &0.750 &0.750 &0.735 &0.446&0.667  \\   
Rolling Bearing   &0.450  &0.500  &0.433  &0.500  &0.933    &0.589  &0.558  &0.500       &0.500 &0.500 &0.500 &0.300 &0.446&0.500  \\ 
Spherical Bearing &0.417  &0.500  &0.883  &0.567  &0.650    &0.682  &0.538  &0.500       &0.500 &0.500 &0.500 &0.450 &0.346&0.500  \\ 
Servo             &0.742  &0.795  &0.769  &0.750  &0.750    &0.277  &0.344  &0.250       &0.250 &0.750 &0.750 &0.675 &0.420&0.237  \\  
Clip              &0.667  &0.619  &0.667  &0.667  &0.667    &0.462  &0.330  &0.330       &0.333 &0.693 &0.667 &0.648 &0.467&0.531  \\  
USB               &0.467  &0.500  &0.508  &0.504  &0.500    &0.530  &0.513  &0.500       &0.496 &0.500 &0.500 &0.563 &0.518&0.500  \\   
Hinge             &0.750  &0.750  &0.750  &0.750  &0.750    &0.686  &0.438  &0.229       &0.283 &0.758 &0.750 &0.250 &0.564&0.250  \\   
Sticky Roller     &0.667  &0.667  &0.667  &0.667  &0.667    &0.694  &0.667  &0.333       &0.333 &0.625 &0.667 &0.556 &0.518&0.333  \\   
Caster Wheel      &0.750  &0.750  &0.533  &0.733  &0.750    &0.765  &0.346  &0.212       &0.250 &0.735 &0.750 &0.650 &0.575&0.250  \\   
Screw             &0.667  &0.667  &0.667  &0.667  &0.667    &0.667  &0.424  &0.333       &0.733 &0.680 &0.667 &0.444 &0.496&0.578  \\   
Lock              &0.633  &0.667  &0.667  &0.667  &0.822    &0.304  &0.333  &0.339       &0.328 &0.667 &0.667 &0.244 &0.470&0.333  \\   
Gear              &0.798  &0.800  &0.738  &0.809  &0.800    &0.244  &0.230  &0.200       &0.209 &0.800 &0.800 &0.780 &0.520&0.793  \\   
Clock             &0.670  &0.670  &0.661  &0.670  &0.670    &0.670  &0.439  &0.330       &0.344 &0.667 &0.670 &0.511 &0.521&0.330  \\   
Slide             &0.800  &0.800  &0.800  &0.800  &0.667    &0.291  &0.410  &0.194       &0.193 &0.800 &0.800 &0.730 &0.494&0.267  \\  
Zipper            &0.661  &0.667  &0.717  &0.667  &0.667    &0.526  &0.345  &0.333       &0.344 &0.667 &0.667 &0.555 &0.508&0.334  \\   
Button            &0.800  &0.800  &0.797  &0.800  &0.800    &0.596  &0.264  &0.197       &0.197 &0.842 &0.800 &0.470 &0.499&0.317  \\   
Liquid            &0.667  &0.667  &0.667  &0.667  &0.667    &0.639  &0.515  &0.333       &0.333 &0.784 &0.667 &0.477 &0.363&0.289  \\  
Rubber Band       &0.500  &0.500  &0.500  &0.500  &0.567    &0.482  &0.519  &0.500       &0.500 &0.500 &0.500 &0.517 &0.515&0.450  \\  
Ball              &0.667  &0.577  &0.667  &0.667  &0.667    &0.667  &0.374  &0.333       &0.333 &0.667 &0.667 &0.570 &0.523&0.644  \\  
Magnet            &0.667  &0.667  &0.689  &0.667  &0.667    &0.548  &0.485  &0.333       &0.333 &0.667 &0.667 &0.600 &0.580&0.489  \\   
Toothpaste        &0.500  &0.500  &0.500  &0.500  &0.500    &0.500  &0.545  &0.500       &0.500 &0.500 &0.500 &0.400 &0.559&0.467  \\  \midrule
Average           &0.658  &0.735  &0.673  &0.669  &0.699    &0.543  &0.417  &0.331       &0.361 &0.673 &0.666 &0.537 &0.493&0.432  \\  \bottomrule
\end{tabular}
}
\label{tab:B}\vspace{-2mm}
\end{table*}

\end{document}